\documentclass[final,5p,times]{elsarticle}

\makeatletter
\def\ps@pprintTitle{%
 \let\@oddhead\@empty
 \let\@evenhead\@empty
 \let\@oddfoot\@empty
 \let\@evenfoot\@empty}
\makeatother

\usepackage{amssymb}
\usepackage{amsmath}

\usepackage{lineno}
\usepackage{textcomp}
\usepackage{stfloats}
\usepackage{url}
\usepackage{verbatim}
\usepackage{graphicx}

\usepackage{subfigure}
\usepackage{subfiles}
\usepackage{amsthm}
\usepackage{booktabs}
\usepackage{xcolor}
\usepackage{rotating}
\usepackage{longtable} 
\usepackage{pdflscape}
\usepackage{array}
\usepackage{enumitem}

\usepackage{tikz}
\usepackage[edges]{forest} 
\usepackage{caption}       

\definecolor{hidden-draw}{RGB}{20,68,106}
\definecolor{hiddendraw}{RGB}{91,155,213}
\definecolor{output-white}{RGB}{255,255,255}
\definecolor{myblue}{RGB}{137,195,235}

\newcommand{\paratitle}[1]{\vspace{1ex}\noindent \textbf{#1}}
\newcommand{\tabincell}[2]{\begin{tabular}{@{}#1@{}}#2\end{tabular}}

\theoremstyle{definition}
\newtheorem{definition}{Definition}[]

\journal{Information Fusion}

\begin{document}

\begin{frontmatter}



\title{Self-Supervised Representation Learning for Geospatial Objects: A Survey}


\author[a]{Yile Chen} 
\affiliation[a]{organization={College of Computing and Data Science, Nanyang Technological University},
            country={Singapore}}

\author[b]{Weiming Huang} 
\affiliation[b]{organization={Department of Physical Geography and Ecosystem Science, Lund University},
            country={Sweden}}

\author[c]{Kaiqi Zhao} 
\affiliation[c]{organization={School of Computer Science, University of Auckland},
            country={New Zealand}}

\author[a]{Yue Jiang} 

\author[a]{Gao Cong} 

\begin{abstract}
The proliferation of various data sources in urban and territorial environments has significantly facilitated the development of geospatial artificial intelligence (GeoAI) across a wide range of geospatial applications. 
However, geospatial data, which is inherently linked to geospatial objects, often exhibits data heterogeneity that necessitates specialized fusion and representation strategies while simultaneously being inherently sparse in labels for downstream tasks. Consequently, there is a growing demand for techniques that can effectively leverage geospatial data without heavy reliance on task-specific labels and model designs. 
This need aligns with the principles of self-supervised learning (SSL), which has garnered increasing attention for its ability to learn effective and generalizable representations directly from data without extensive labeled supervision. This paper presents a comprehensive and up-to-date survey of SSL techniques specifically applied to or developed for geospatial objects in three primary vector geometric types: \textit{Point}, \textit{Polyline}, and \textit{Polygon}. We systematically categorize various SSL techniques into  predictive and contrastive methods, and analyze their adaptation to different data types for representation learning across various downstream tasks. Furthermore, we examine the emerging trends in SSL for geospatial objects, particularly the gradual advancements towards geospatial foundation models.  Finally, we discuss key challenges in current research and outline promising directions for future investigation. By offering a structured analysis of existing studies, this paper aims to inspire continued progress in integrating SSL with geospatial objects, and the development of geospatial foundation models in a longer term.
\end{abstract}




\end{frontmatter}


\section{Introduction}\label{sec:intro}

The digitization of urban and territorial environments has significantly expanded the collection of extensive geospatial data associated with various objects on our planet, including road segments, buildings, neighborhoods, etc. The vast repository of data serves as the foundation of smart city applications, such as spatial keyword search~\cite{VLDBJ_spatialquery,spatialquery}, location-based services~\cite{VLDB17_poi,DeepJMT}, geospatial knowledge graph~\cite{SIGMOD23_geokg}, intelligent transportation systems~\cite{traffic_survey,miao2024unified}, and socioeconomic indicator prediction~\cite{socio1,socio2}. Despite the richness of geospatial data, its effective mining and utilization remain challenging. In particular, a fundamental limitation arises from the task-specific nature of many deep learning models. These models are typically trained using supervised learning in specific tasks with abundant domain-specific labeled datasets (e.g., traffic records), which can be limited or costly to acquire and restricted by data privacy regulations~\cite{gdpr}. Additionally, while various urban tasks exhibit intrinsic commonalities, such as the close relationship between population density and land use patterns, these models usually suffer from limited generalization across downstream applications due to their reliance on task-specific supervision signals and specialized model architectures. This lack of adaptability restricts the broader applicability of previous approaches across different geospatial analytics tasks, which entails the need for more flexible and generalizable learning paradigms for a variety of geospatial analyses.

In response to these challenges, self-supervised learning \allowbreak(SSL) ~\cite{SSL_survey} has emerged in recent years as a promising paradigm that reduces the dependency on annotated labels while producing task-agnostic and general-purpose data representations. The core principle of SSL is to extract transferrable knowledge from the target data through well-designed self-supervised tasks (i.e., pretext tasks), wherein the supervision signals are automatically generated from the data itself. SSL has achieved notable success across various domains and diverse data modalities, including images~\cite{SSL_image1,SSL_image2}, videos~\cite{SSL_video}, language~\cite{SSL_text,SSL_text2}, graphs~\cite{SSL_graph1,SSL_graph2}, time series~\cite{SSL_timeseries}, etc. For example, massive text corpora are structured in an autoregressive generation framework, which is well-suited for next token prediction for the training of large language models (LLMs)~\cite{GPT3}. In addition, images are processed through data augmentation operations to produce multiple views, with models trained to produce invariant representations across views for computer vision tasks~\cite{SSL_image1}.

A key motivation for applying SSL techniques in the geospatial domain is to learn effective and generalizable representations (embeddings) for various forms of geospatial objects, such as points-of-interest (POI), road segments, and urban regions. These objects underpin a variety of human activities within urban environments, and therefore serve as fundamental analytical units for numerous urban analytical applications. For example, road networks are critical infrastructures that support human movement activities within a city. As a result, a variety of tasks, such as the prediction of traffic speed, congestion, and destination, commonly regard individual road segments as analytical units. In these scenarios, SSL offers a powerful framework for deriving general-purpose representations of geospatial objects through fusing information from multiple perspectives, capturing both the intrinsic characteristics of each object and the complex interplay among different objects. The representations learned through SSL can be readily leveraged to train simpler models (e.g., a linear model) for various downstream tasks while maintaining effective performance.

Apart from the strong generalization capabilities, the interest in applying SSL techniques is also driven by the ability to work without extensive labeled datasets. This advantage addresses a long-standing challenge in the geospatial domain-- the scarcity of  task-specific labeled data. As a result, SSL presents a viable alternative to the conventional supervised learning \allowbreak paradigm, which necessitates the development of specialized deep learning models trained on sufficiently labeled data for each downstream application. By leveraging the self-supervised signals derived from data itself, SSL has the potential to better tackle diverse geospatial (urban) analytical applications.

However, geospatial objects, embedded within geographic spaces, exhibit various forms and spatial attributes while adhering to geographical principles, such as the Laws of Geography~\cite{geolaw}. These spatial characteristics introduce significant challenges when applying standard SSL techniques developed for other domains, as they often fail to capture the intricate spatial semantics attached to geospatial objects. Besides, certain SSL components, such as data augmentation and pretext tasks, must be carefully designed and adapted to preserve the spatial and structural integrity of geospatial objects. Unlike in vision or language domains, where augmentations such as cropping, rotation, or synonym replacement are widely used, augmentations for spatial objects must ensure consistency with geographical relationships and dependencies. Moreover, geospatial objects are often associated with heterogeneous context information, capturing diverse yet complementary aspects of their intrinsic properties. Integrating this information requires the development of effective modeling and fusion strategies to produce high-quality data representations.  Considering these unique challenges in the geospatial domain, recent research has introduced novel SSL techniques that incorporate spatial awareness and domain-specific constraints, ensuring the effective adaptation of SSL within the geospatial context.

Despite the growing body of literature, the application of SSL within the geospatial domain remains insufficiently discussed and summarized. To bridge this gap, this survey provides a comprehensive and systematic review of up-to-date SSL techniques tailored or developed for learning geospatial object representations, which in turn facilitate various geospatial analyses. In particular, we focus on three primary geospatial data types, categorized based on their geometric forms: points, polylines, and polygons. We adopt a structured framework to present the specialized SSL studies for these data types, focusing particularly on methods that operate independently of specific tasks and supervised settings.  For each data type, we analyze how SSL techniques encode intrinsic attributes while integrating heterogeneous context information associated with geospatial objects. Besides, we review studies that utilize SSL to fuse multiple geospatial data types for learning representations. Building on the insights from the surveyed studies, we identify and analyze emerging trends in the development of geospatial foundation models.
Furthermore, since SSL techniques serve as a universal paradigm that can also be utilized to enhance the supervised models as auxiliary objects in geospatial applications such as spatio-temporal forecasting, we provide a brief overview of task-specific SSL implementations applied to several domain-specific scenarios, supplementing the research landscape of SSL in this field.  
This survey aims to cover a wide range of model scopes, from specialized SSL models to advanced foundation models, applied to different geospatial data types and analyses. The main contributions of this survey are summarized as follows:

\begin{itemize}
    \item We present a detailed and up-to-date review of SSL techniques for  geospatial objects, focusing on three types of geospatial data, mainly in urban environments: \textit{Point}, \textit{Polyline}, and \textit{Polygon (Region)}. To the best of our knowledge, this work is the first to systematically discuss SSL techniques tailored for learning representations in the \allowbreak geospatial domain.
    \item We introduce a comprehensive and structured way for specialized SSL models designed for the studied data types. Our categorization includes an analysis of intrinsic characteristics and heterogeneous context information within each data type, encompassing discussion on both predictive and contrastive SSL implementations. 
    \item We review several recent advancements based on foundation models and task-specific SSL techniques for geospatial data, providing insights into the emerging trends. 
    \item We discuss several key challenges for SSL in geospatial domain, and propose potential future directions to advance the related research.  
\end{itemize}

\textit{Related Surveys and Our Distinction:} Several recent surveys have discussed the application of SSL, mainly focusing on other domains, such as general SSL~\cite{SSL_survey}, SSL for computer vision~\cite{SSL_image2}, graphs~\cite{SSL_graph1, SSL_graph2}, time series~\cite{SSL_timeseries}, and recommender systems~\cite{SSL_recsys}.  However, despite SSL’s growing importance in data-driven analytics, its adaptation in the geospatial domain remains underexplored. Recognizing this limitation, our survey presents a comprehensive and systematic review of SSL tailored for geospatial data, which serves as the foundations in numerous downstream applications to enhance urban intelligence. On the other hand, several recent surveys have paid attention to spatio-temporal data analytics with varying emphases, such as trajectory data mining~\cite{traj_survey1, traj_survey2}, urban foundation models~\cite{urban_survey}, geo-location encoding~\cite{survey_geoencoding}, and supervised or generative deep learning~\cite{st_survey, st_survey2}. While these works contribute to the broader field of GeoAI and may overlap with some studies discussed in this paper, 
our survey distinguishes itself by structuring knowledge from the unique perspective of SSL. Specifically, we provide a detailed summary of SSL developments across different geospatial data types, systematically categorizing existing methodologies and synthesizing their implementation in geospatial contexts.

\textit{Paper Structure:} The rest of this survey is organized as follows. Section~\ref{sec:preliminary} provides definitions, preliminary concepts, and background knowledge necessary for the subsequent sections. Section~\ref{sec:points},~\ref{sec:polylines}, and~\ref{sec:polygons} look into the details of specialized SSL techniques applied to data with three distinct geometric types: points, polylines and polygons, respectively. For each data type, representative data instances are further elaborated in the context of SSL, including POI for points, trajectory and road network for polylines, and region for polygons. Section~\ref{sec:multi-type} presents SSL techniques for the integration of multiple data types. Section~\ref{sec:trending} presents our discussion on the emerging trends  based on the development of geospatial foundation models, and provides an overview for task-specific SSL techniques for geospatial data. Section~\ref{sec:discussion} provides several promising future research directions. Finally, Section~\ref{sec:conclusion} concludes this paper. 

\section{Definition and Background}\label{sec:preliminary}
In this section, we introduce the definition of three types of geospatial vector data examined in this paper. Then we present the paradigms of SSL applied to geospatial objects based on the traits of pretext tasks. Last, we discuss preliminaries on typical models that encode these geospatial data types.

\subsection{Definition of Geospatial Data Types}
\begin{figure}[t]
  \centering
  \subfigure[Points] {
        \begin{minipage}[t]{1.0\linewidth}
        \centering
        \includegraphics[width=3.2in]{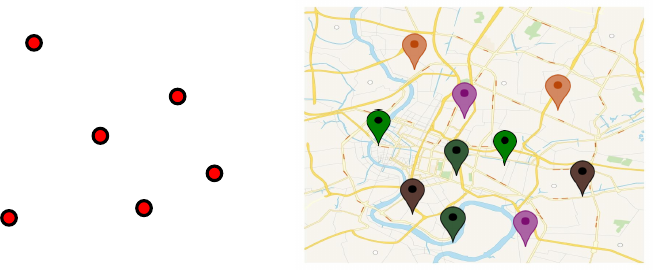}
        \end{minipage}
        \label{subfig:point}
    }
    \subfigure[Polylines] {
        \begin{minipage}[t]{1.0\linewidth}
        \centering
        \includegraphics[width=3.2in]{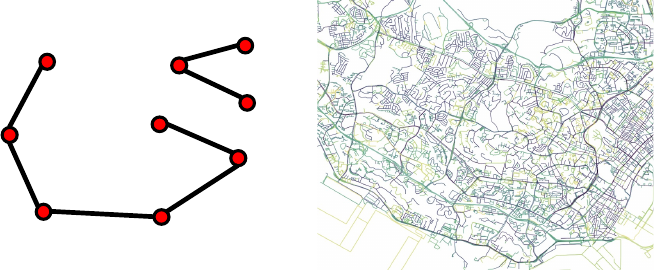}
        \end{minipage}
        \label{subfig:polyline}
    }
    \subfigure[Polygons] {
        \begin{minipage}[t]{1.0\linewidth}
        \centering
        \includegraphics[width=3.2in]{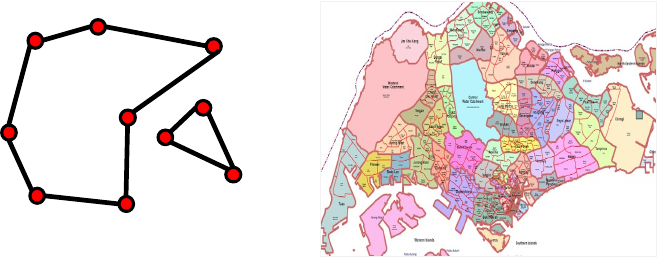}
        \end{minipage}
        \label{subfig:polygon}
    }
\vspace{-3mm}
\caption{Three types of geospatial objects and their data instances.}
\label{fig:spatial_objects}
\vspace{-4mm}
\end{figure}

In this survey, we adopt the widely recognized classification scheme in spatial database research, which categorizes geospatial objects by their geometric representations into three types: \textit{Point}, \textit{Polyline}, and \textit{Polygon (Region)}~\cite{guting1994introduction}. This scheme is well-suited to our study, as each type corresponds to distinct real-world geospatial phenomena that require different representation strategies. The three data types are illustrated in Figure~\ref{fig:spatial_objects}. The formal definitions of the three data types are provided as follows.

\begin{definition}[\textbf{Point}]
A geospatial point object is represented as $p = (l, x)$, where $l$ denotes the geographical coordinates, and $x$ refers to the associated features of the point, such as attributes or readings. This data type indicates the spatial locations equipped with contextual information, applicable to data instances such as POIs and sensor measurements.
\end{definition}

\begin{definition}[\textbf{Polyline}]
A geospatial polyline object is defined as a sequence of connected line segments, represented by a list of vertices $\mathcal{L}= [(l_1, x_1), ..., (l_n, x_n)]$, where $l_i$ denotes the geographical coordinates of the $i$-th point, and $x_i$ denotes its associated features, such as timestamps or semantic tags. This data type captures the sequential nature and directionality of spatial paths, applicable to data instances such as trajectories and road networks.
\end{definition}

\begin{definition}[\textbf{Polygon}]
A geospatial polygon object is defined as a closed shape (region) consisting of a sequence of line segments that connect to enclose an area, denoted by $\mathcal{B} = [(l_1, l_2), ..., (l_{n-1}, l_{n}), (l_n, l_1)]$, where $l_i$ represents the geographical coordinates of the $i$-th vertex. The vertices are connected sequentially, with the first vertex repeated as the last to close the polygon. By default, the line segments of a polygon are arranged to ensure that it does not intersect with itself, maintaining a simple, closed loop. This data type can be employed to describe administrative regions or subzones in urban spaces. Therefore, research on polygons often extends its focus beyond its geometric configuration, placing emphasis on the semantic patterns exhibited by objects within or close to its enclosed area.
\end{definition}

\subsection{Paradigms for Self-supervised Learning}
Geospatial data instances in practical scenarios can be conceptualized as a combination of geometric forms with intrinsic attributes (e.g., spatial coordinates) and associated context information (e.g., textual content).  
Therefore, SSL process in geospatial context aims to integrate the geometric form of the target geospatial object and its contextual information to derive effective representations. These representations are learned through the training objectives (pretext tasks), where supervision signals are automatically extracted from the geospatial data itself, eliminating the need for additional label annotations.
\allowbreak Based on the design of pretext tasks, SSL techniques for geospatial objects can be divided into two categories: predictive and contrastive methods.

\subsubsection{Predictive Methods}
Predictive methods employ pretext tasks that are formulated as prediction problems,  with objectives derived from the original data instances. Specifically, these methods involve tasks like the reconstruction of corrupted geospatial objects using a subset of available data, or the prediction of auxiliary labels that are readily extracted from the attributes or structures of geospatial objects. They can be formulated as:

\begin{equation}
    f_{\theta}^{*}, p_{\phi}^{*}=\underset{f_{\theta}, p_{\phi}}{\arg \min } \mathcal{L}_{pre}\left(p_\phi\left(f_\theta(\mathcal{D}, \mathcal{D}_{c})\right), \mathcal{D}_t\right)
\end{equation}
where $f_\theta$ and $p_{\phi}$ represent the geospatial encoder and the pretext decoder respectively. The geospatial encoder, which is introduced in Section~\ref{subsec:encoder}, is responsible for deriving representations for geospatial objects, while the pretext decoder is typically a lightweight structure, such as a shallow multi-layer perceptron (MLP), to map the representations to the space of the prediction objectives. $\mathcal{D}$ denotes the target geospatial objects, which can be of any data types. $\mathcal{D}_{c}$ is the context information associated with $\mathcal{D}$, and $\mathcal{D}_{t}$ denotes the prediction objectives, which could either be the original data instance for reconstruction tasks, or additional features excluded from the encoder input for auxiliary label prediction. $\mathcal{L}_{pre}$ is the loss function that measures the prediction error, such as cross-entropy loss or mean squared error (MSE) loss.

\subsubsection{Contrastive Methods}
Contrastive methods are based on the principle of maximizing agreement between different views generated from the same data instance. Specifically, these methods aim to pull closer the representations of positive view pairs, which are derived from various data augmentation operations of the same data instance, while pushing apart the representations of negative view pairs from different data instances. They can be formulated as:

\begin{equation}
    f_{\theta}^{*}, p_{\phi}^{*}=\underset{f_{\theta}, p_{\phi}}{\arg \min } \mathcal{L}_{con}\left(p_\phi\left(f_\theta(\mathcal{\Tilde{D}}^{1}, \mathcal{\Tilde{D}}_{c}^{1})\right), p_\phi\left(f_\theta(\mathcal{\Tilde{D}}^{2}, \mathcal{\Tilde{D}}_{c}^{2})\right) \right)
\end{equation}
where $f_\theta$ and $p_{\phi}$ represent the geospatial encoder and the pretext decoder respectively. The pretext decoder is usually a projection head for linear transformation~\cite{SSL_image1}. $\mathcal{\Tilde{D}}^{1}$ and $\mathcal{\Tilde{D}}^{2}$ are two distinct views generated from the target geospatial objects $\mathcal{D}$, which can belong to any geospatial data types, and $\mathcal{\Tilde{D}}_{c}^{1}$ and $\mathcal{\Tilde{D}}_{c}^{2}$ are the context information associated with these respective views. $\mathcal{L}_{con}$ is the contrastive loss function that quantifies the degree of agreement, typically measured by mutual information estimator~\cite{SSL_survey}, such as InfoNCE~\cite{InfoNCE}, JS divergence~\cite{JS} and triplet loss~\cite{Triplet}.

\subsection{Preliminaries on Geospatial Encoder}\label{subsec:encoder}
Given the diversity of geospatial data types discussed in this survey, each with its unique geometric (locational) and intrinsic properties, the utilized geospatial encoder $f_\theta$  would be varied to accommodate their distinct characteristics. Therefore, we provide a brief introduction on several neural network modules frequently employed or adapted as geospatial encoders. 

\subsubsection{Graph Neural Networks}
Graph Neural Networks (GNNs)~\cite{GNN_survey} correspond to a type of neural network architectures designed to process graph-struc \allowbreak tured data, aiming to capture the complex relationships and structures within the graph. GNNs employ message-passing operations iteratively on the graph, where the representation of a node $v$ is updated through interactions with its neighbors. This process can be expressed as:

\begin{equation}
h_v^{(l)}  =\mathcal{F}^{(l)}\left(h_v^{(l-1)}, \operatorname{AGG}^{(l)}\left(\left\{h_u^{(l-1)}\right\}_{u \in \mathcal{N}_v}\right)\right)  
\end{equation}
where $h_v^{(l)}$ indicates the representation of $v$ at layer $l$ and $\mathcal{N}_v$ denotes the neighbors of $v$. $\operatorname{AGG}^{(l)}$ is the message aggregation function at layer $l$, which collects and combines node features, and potentially edge features, from the neighbors, and $\mathcal{F}^{(l)}$ is the function that updates the representation of $v$ based on the aggregated information. For geospatial objects, GNNs are frequently utilized to model discrete objects, enabling the capture of complex relationships among them.

\subsubsection{Sequential Models}
Sequential models are designed to process input data composed of sequences, which include domains such as time series, text, audio, and video. Over the past decade, neural network architectures have exhibited exceptional performance in sequence modeling due to their capability of capturing dependencies effectively. 
The general process can be described as:
\begin{equation}
[h_1, ..., h_n]  = \mathcal{F}\left([x_1, ..., x_n]; \Theta \right)  
\end{equation}
where $[x_1, ..., x_n]$ denotes the input sequence, $[h_1, ..., h_n]$ denotes the hidden representations output by the sequential model $\mathcal{F}$, which is parameterized by $\Theta$. 
Recurrent Neural Networks (RNNs) accomplish this by recursively processing the current input along with previous elements of the sequence, where the previous elements are encoded into internal hidden states, leading to several model variants, with GRU~\cite{GRU} and LSTM~\cite{LSTM} being the most notable ones. 
In recent years, the Transformer architecture~\cite{Transformer} has revolutionized sequence modeling by handling historical sequences in a parallel manner, instead of the recursive approaches. Meanwhile, it demonstrates superiority of modeling pairwise relationships between any two positions in a sequence through self-attention mechanism.
For geospatial objects, sequential models are  particularly valuable for modeling trajectories or data instances that are built to consider the sequential dependencies.

\subsubsection{Pre-trained Models}
The evolution of advanced sequential models, especially those based on the Transformer architecture, have marked the milestone in the development of pre-trained language models. One popular paradigm is to adhere to the principles set by BERT\allowbreak ~\cite{BERT}. These models~\cite{ALBERT, RoBERTa, BART, T5} leverage large-scale datasets to employ the training objective of masked language modeling (MLM).  This process enables the acquisition of rich and transferable representations, which can be fine-tuned for specific tasks with less labeled data. Another paradigm is to employ training with autoregressive language modeling, exemplified by ChatGPT and its counterparts~\cite{InstructGPT, GPT4, Gemini, Llama2, Claude}. These large language models (LLMs) demonstrate strong capabilities in language understanding and reasoning, transforming various tasks into the process of autoregressive language generation. The robust performance of these two paradigms has led to the widespread use of pre-trained language models to encode textual information associated with geospatial data or to adapt their training objectives to develop specialized representations. On the other hand, in scenarios where vision information, such as street view images, is associated with geospatial elements like polylines or polygons, pre-trained visual models are also utilized. These include CNN-based models~\cite{Resnet, Mobilenet} and recent Transformer-based models~\cite{ViT, SwinViT} trained on large-scale image datasets. When both textual and visual data sources are available, visual-language pre-trained models such as CLIP and its variants~\cite{CLIP, BLIP}  are employed to synergize the semantics of the two data modalities, enhancing the interpretability and utility of combined data sources in geospatial applications.  

\begin{figure}[tbp]
    \centering
    \vspace{-2pt}
    \includegraphics[width=0.98\linewidth]{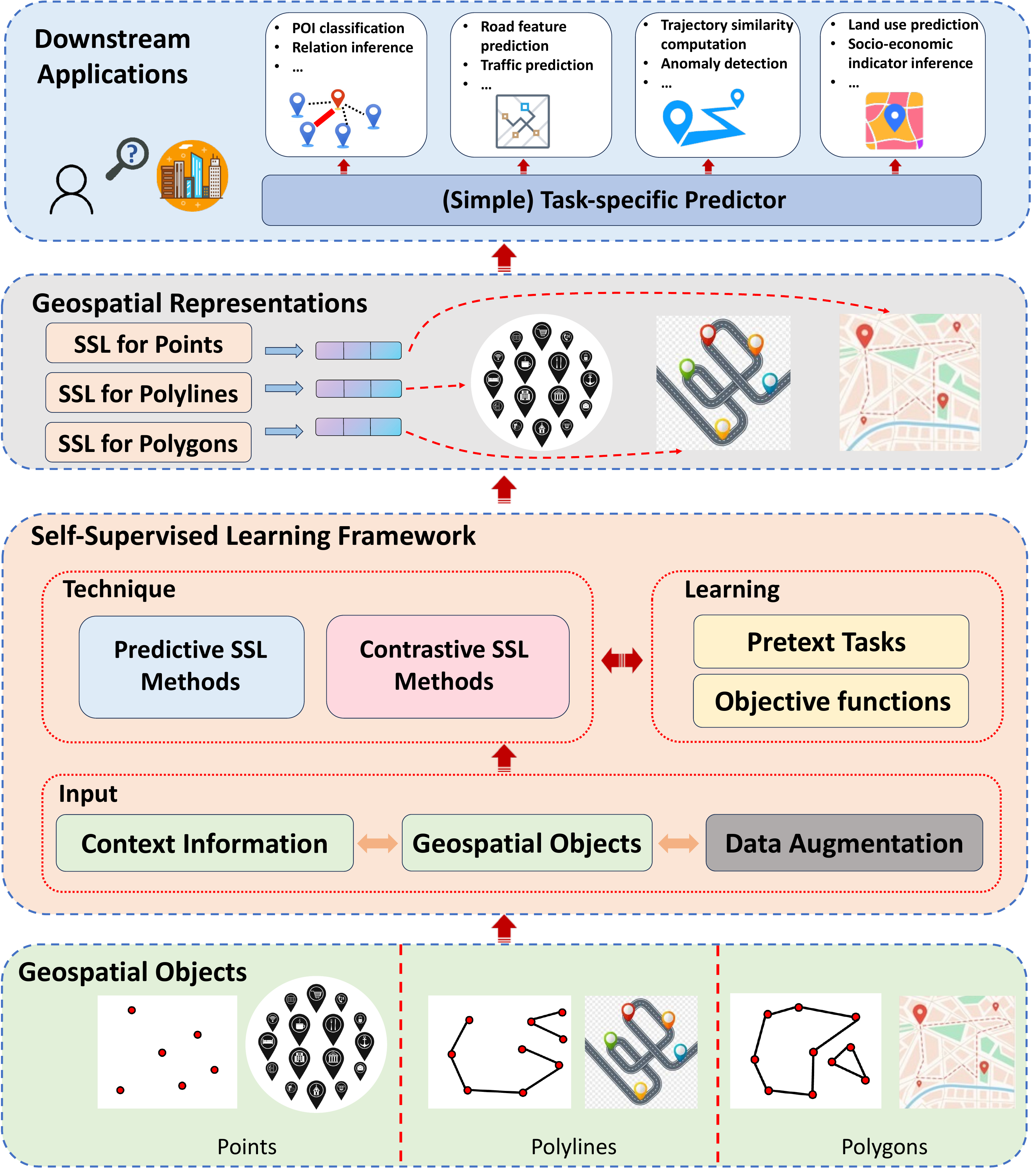}
    \caption{Overview of SSL framework for geospatial objects.}
    \label{fig:framework}
    \vspace{-4mm}
\end{figure}

\begin{figure}[h]
  \centering
  \subfigure[Predictive SSL methods] {
        \begin{minipage}[t]{1\linewidth}
        \centering
        \includegraphics[width=\linewidth]{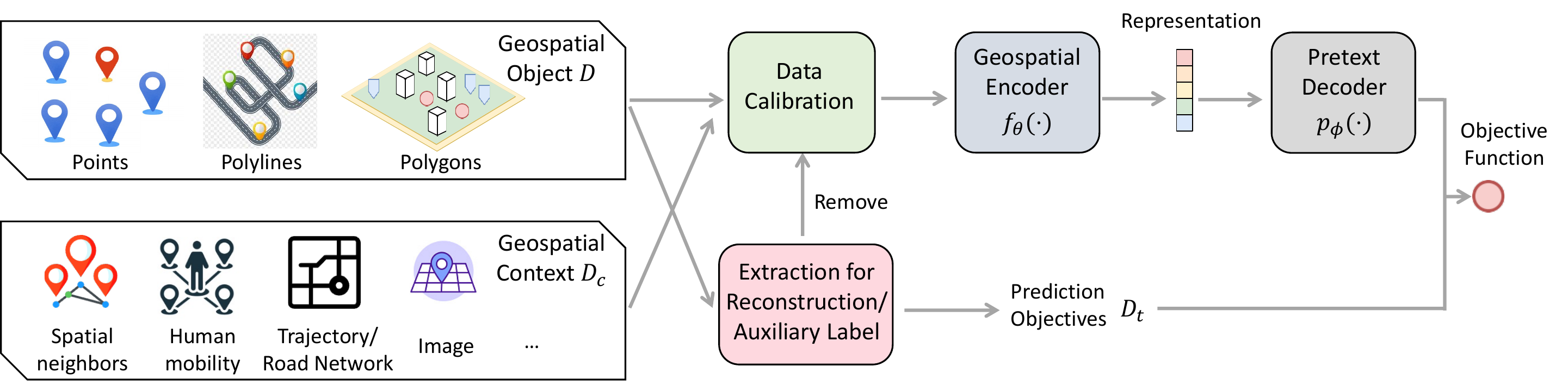}
        \end{minipage}
        \label{subfig:predictive}
    }
    \subfigure[Contrastive SSL methods] {
        \begin{minipage}[t]{1\linewidth}
        \centering
        \includegraphics[width=\linewidth]{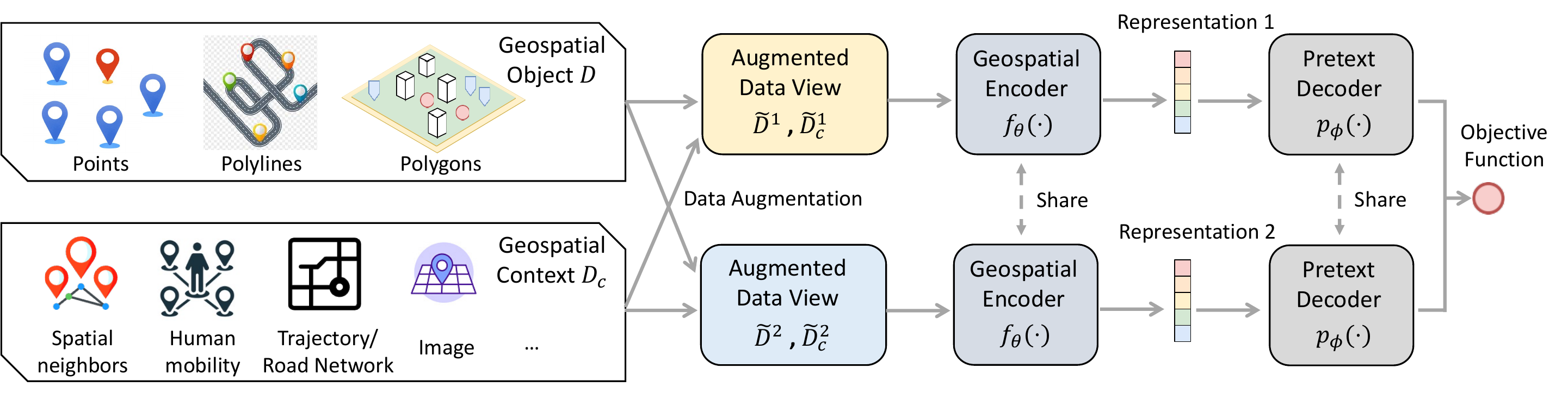}
        \end{minipage}
        \label{subfig:contrastive}
    }
\caption{Illustration of two types of SSL methods for geospatial objects.}
\label{fig:ssl_type}
\end{figure}

\tikzstyle{my-box}=[
    rectangle,
    draw=hidden-draw,
    rounded corners,
    align=left,
    text opacity=1,
    minimum height=1.5em,
    minimum width=5em,
    inner sep=2pt,
    fill opacity=.8,
    line width=0.8pt,
]

\tikzstyle{leaf-head}=[my-box, minimum height=1.5em,
    draw=yellow!80, 
    fill=yellow!35,  
    text=black, font=\normalsize,
    inner xsep=2pt,
    inner ysep=4pt,
    line width=0.8pt,
]

\tikzstyle{leaf}=[draw=hiddendraw,
    rounded corners, 
    minimum height=1.8em,
    fill=myblue!40,text opacity=1, 
    fill opacity=.5,  text=black,align=left,
    font = \normalsize,
    inner xsep=2pt,
    inner ysep=2pt,
    ]
    
\tikzstyle{middle}=[draw=hiddendraw,
    rounded corners,
    minimum height=1.8em,
    fill=output-white!40, text opacity=1, 
    fill opacity=.5,  text=black, align=center,
    font = \normalsize,
    inner xsep=2.5pt,
    inner ysep=2pt,
    fill opacity=.8,
    ]

\begin{figure*}[!th]
    \centering
    \resizebox{0.9\textwidth}{!}{
        \begin{forest}
            forked edges,
            for tree={
                grow=east,
                reversed=true,
                anchor=base west,
                parent anchor=east,
                child anchor=west,
                base=left,
                font=\normalsize,
                rectangle,
                draw=hidden-draw,
                rounded corners,
                align=left,
                minimum width=1em,
                edge+={darkgray, line width=1pt},
                s sep=10pt,
                inner xsep=0pt,
                inner ysep=3pt,
                line width=0.8pt,
                ver/.style={rotate=90, child anchor=north, parent anchor=south, anchor=center},
            }, 
            [
               SSL for Different Geospatial Data Instances, leaf-head, ver
                [
                    {POIs}, middle,text width=8em
                    [
                        Spatial Neighborhood, middle, text width=10.5em,
                         [{\textbf{Predictive}: Place2vec~\cite{place2vec}, Semantic~\cite{semantic_poi}, MT-POI~\cite{mt_poi}, STPA~\cite{poi-annotate-ijcai23},\\ 
                         DeepR~\cite{POI-competitive-kdd20}, PRIM~\cite{POI-relation-vldb22}, GeoBERT\cite{geobert}, SpaBERT~\cite{spabert}\\
                         \textbf{Predictive\&Contrastive}: MGeo~\cite{mgeo}}, leaf, text width=38em
                        ]    
                    ]
                    [
                        Check-in Sequence, middle, text width=10.5em
                         [{\textbf{Predictive}: SG-CWARP~\cite{poi-context-ijcai16}, CAPE~\cite{poi-hier-ijcai18}, DeepMove~\cite{deepmove}, Hier-CEM~\cite{hier-embed},\\
                         POI2Vec~\cite{poi2vec}, TALE~\cite{TALE}, LSPSL~\cite{LSPSL}
                         }, leaf, text width=38em
                        ]    
                    ]
                    [
                        Co-query Context, middle, text width=10.5em
                       [{\textbf{Contrastive}: MoCo-GA~\cite{MoCo-GA}
                         }, leaf, text width=38em
                        ] 
                    ]
                    [
                        Multiple Sources, middle, text width=10.5em
                        [{\textbf{Predictive}: CatEM~\cite{CatEM}, ERNIE-GeoL~\cite{ERNIE-GeoL}\\
                        \textbf{Contrastive}: SEENet~\cite{POI-relation-kdd23}, DCHL~\cite{DCHL}, POI-Enhancer~\cite{poienhancer}\\
                        \textbf{Predictive\&Contrastive}: STKG-PLM~\cite{STKG-PLM}, AGCL~\cite{AGCL}
                         }, leaf, text width=38em]
                    ]
                ]
                [
                    {Road Networks}, middle,text width=8em
                    [
                        Intrinsic Attributes, middle, text width=10.5em,
                        [   
                        {\textbf{Predictive}: R2Vec~\cite{road_BigData18}, IRN2Vec~\cite{road_IRN2Vec}, RFN~\cite{road_RFN}, HyperRoad~\cite{road_HyperRoad}}, leaf, text width=38em
                        ]    
                    ]
                    [
                        Spatial Attributes, middle, text width=10.5em,
                        [{\textbf{Predictive}: RN2Vec~\cite{road_RN2Vec} \;\
                        \textbf{Contrastive}: SARN~\cite{road_SARN}}, leaf, text width=38em
                        ]    
                    ]
                    [
                        Trajectories, middle, text width=10.5em
                         [{\textbf{Predictive}: HNRN~\cite{road_HNRN}, Toast~\cite{road_Toast}, DyToast~\cite{Dytoast}, TrajRNE~\cite{road_TrajRNE} \;\  \\
                        \textbf{Contrastive}: JCLRNT~\cite{road_JCLRNT} \;\ \textbf{Predictive\&Contrastive}: TRACK~\cite{road_track}}, leaf, text width=38em
                        ]    
                    ]
                    [
                        Street View Images, middle, text width=10.5em
                       [{\textbf{Contrastive}: Garner~\cite{road_Garner} \;\  
                        \textbf{Predictive\&Contrastive}: USPM~\cite{road_KDD24}}, leaf, text width=38em
                        ]  
                    ]
                ]
                [
                    Trajectory, middle,text width=8em
                        [
                        Intrinsic Attributes, middle, text width=10.5em,
                        [{\textbf{Predictive}: TCDRL~\cite{traj_IJCNN17}, t2vec~\cite{traj_t2vec}, GM-VSAE~\cite{traj_ICDE20}, Geo-Tokenizer~\cite{traj_ECML23}, \\ AdvTraj2vec~\cite{traj_CIKM22_adv}, E$^{2}$DTC~\cite{traj_ICDE21}, STPT~\cite{traj_ICDM23}, ST-t2vec~\cite{traj_TIST21}, DeepTEA~\cite{traj_VLDB22}, UniTraj~\cite{unitraj}\\ 
                        \textbf{Contrastive}: CL-Tsim~\cite{traj_CIKM22_contrastive}, TrajCL~\cite{traj_ICDE23_contrastive}, KGTS~\cite{traj_AAAI24_contrastive} \\
                        \textbf{Predictive\&Contrastive}: RSTS~\cite{traj_WWWJ23}}, leaf, text width=38em
                        ]  
                    ]
                    [
                        Road Networks, middle, text width=10.5em,
                        [{\textbf{Predictive}: CSSRNN~\cite{traj_IJCAI17_road}, Traj2Vec~\cite{traj_TIST_road}, PT2Vec~\cite{traj_DASFAA23_road}, JGRM~\cite{traj_WWW24_road}\\ 
                        \textbf{Contrastive}: ST2Vec~\cite{traj_KDD22_road}, GRLSTM~\cite{traj_AAAI23_road}, PIM~\cite{traj_IJCAI21_road}, MMTEC~\cite{traj_TKDE_road} \\
                        \textbf{Predictive\&Contrastive}: HMTRL~\cite{traj_VLDBJ_road}, START~\cite{traj_ICDE23_road}, LightPath~\cite{traj_KDD23_road},\\ GREEN~\cite{traj_KDD25_road}}, leaf, text width=38em
                        ]     
                    ]
                    [
                        Semantic Information, middle, text width=10.5em
                         [{\textbf{Predictive}: CTLE~\cite{traj_AAAI21_semantic}, At2vec~\cite{traj_TKDE23_semantic} \\ 
                        \textbf{Contrastive}: SML-LP~\cite{traj_KBS_semantic}, CLMTR~\cite{traj_clmtr}, CACSR~\cite{traj_AAAI23_semantic} }, leaf, text width=38em
                        ]    
                    ]
                    [
                        Multiple Sources, middle, text width=10.5em
                       [{\textbf{Predictive}: RED~\cite{traj_red} \;\
                        \textbf{Contrastive}: PTrajM~\cite{traj_ptrajm}\\ \textbf{Predictive\&Contrastive}: MVTraj~\cite{traj_mvtraj}, MM-Path~\cite{traj_KDD25_mmpath}}, leaf, text width=38em
                        ]    
                    ]
                ]
                [
                    Urban Regions, middle,text width=8em
                    [
                        Intrinsic Attributes, middle, text width=10.5em,
                        [{\textbf{Contrastive}: Triplet~\cite{liu2018efficient}, UrbanVLP~\cite{hao2024urbanvlp}, MMGR~\cite{bai2023geographic}\\ \textbf{Predictive\&Contrastive}: UrbanCLIP~\cite{yan2024urbanclip}}, leaf, text width=38em
                        ] 
                    ]
                    [
                        POIs/ Knowledge\\ \;\;\;\ Graphs, middle, text width=10.5em,
                        [{\textbf{Contrastive}: HGI~\cite{huang2023hgi} \;\ \textbf{Predictive\&Contrastive}: SKRL4RS~\cite{jin2019learning}}, leaf, text width=38em
                        ] 
                    ]
                    [
                        Human Trajectories, middle, text width=10.5em
                         [{\textbf{Predictive}: HDGE~\cite{wang2017region}, ZE-Mob~\cite{yao2018representing}, GMEL~\cite{liu2020learning}, MGFN~\cite{wu2022multi}}, leaf, text width=38em
                        ] 
                    ]
                    [
                        POIs + Human \\Trajectories (+ others), middle, text width=10.5em
                       [{\textbf{Predictive}: MP-VN~\cite{fu2019efficient}, DLCL~\cite{du2019beyond}, CGAL~\cite{zhang2019unifying}, HAFusion~\cite{sun2024urban}  \\
                       MVURE~\cite{zhang2021multi}, Region2Vec\cite{luo2022urban}, HUGAT\cite{kim2202effective} \\
                         \textbf{Contrastive}: ReMVC~\cite{zhang2022region} \\   \textbf{Predictive\&Contrastive}: HREP~\cite{zhou2023heterogeneous}, ReCP~\cite{li2024urban}}, leaf, text width=38em
                        ] 
                    ]
                    [
                        Imagery Data, middle, text width=10.5em
                       [{\textbf{Predictive}: RegionEncoder~\cite{jenkins2019unsupervised} \\
                       \textbf{Contrastive}: Tile2Vec~\cite{jean2019tile2vec}, Urban2Vec~\cite{wang2020urban2vec}, M3G~\cite{huang2021learning}, Xi et al~\cite{xi2022beyond}, \\ GeoHG~\cite{zou2024learning}, MuseCL~\cite{musecl} \\
                         \textbf{Predictive\&Contrastive}: ReFound~\cite{refound}}, leaf, text width=38em
                        ] 
                    ]
                    [
                        OpenStreetMap, middle, text width=10.5em
                       [{
                       \textbf{Contrastive}: RegionDCL~\cite{li2023urban}}, leaf, text width=38em
                        ] 
                    ]
                ]
            ]
        \end{forest}
    }
    \caption{Taxonomy of SSL methods for geospatial data instances discussed in the paper. It is categorized by the type of context information leveraged to for encoding each data instance.}
    \label{fig:taxonomy}
\end{figure*}

It is important to note that these representative models are not mutually exclusive, and they can be employed concurrently to encode geospatial objects through model combination or to incorporate diverse contextual information.  In the following sections, we adopt a structured analytical approach for each geospatial data type under consideration.  We first present the encoding methods for intrinsic attributes that reflect the fundamental characteristics of the data instance. Then we discuss methods to integrate and fuse heterogeneous context information to enhance the extraction of meaningful insights and knowledge for the studied geospatial object. Finally, we introduce the downstream applications that benefit from the derived geospatial representations.

\subsection{Overview and Taxonomy}

Figure~\ref{fig:framework} provides an overview of the Self-Supervised Learning (SSL) framework for geospatial objects, illustrating the \allowbreak pipeline from raw data to downstream applications. At the foundation, geospatial objects are categorized into three primary geometric data types--points, polylines, and polygons--each corresponding to different data instances such as POIs, road segments, trajectories, and urban regions. These objects and their context information and augmented variations serve as input to the SSL framework. Within this framework, predictive methods focus on recovering missing or masked attributes, while contrastive methods aim to learn representations by distinguishing between similar and dissimilar instances. These techniques are guided by specific pretext tasks and objective functions, such as the cross-entropy for attribute classification, or the InfoNCE loss for contrastive learning. Lightweight task-specific models then utilize the learned representations to support various geospatial applications, such as POI classification, traffic prediction, anomaly detection, socio-economic inference, etc.

The taxonomy of SSL methods discussed in the paper is presented in Figure~\ref{fig:taxonomy}. It is organized according to data instance and the contextual information exploited for representation learning.
The taxonomy comprises four types of geospatial data instances, each grouped according to the specific context information they utilize. We 
classify SSL methods within each branch into predictive, contrastive, or hybrid (predictive \& contrastive) methods. To enhance conceptual clarity, Figure~\ref{fig:ssl_type} provides a schematic illustration for predictive and contrastive SSL methods, as introduced in Section 2.2, applied across different geospatial objects. This illustration outlines the core modeling workflows and highlights the general design principles underlying each SSL paradigm. Building on these paradigms, each geospatial data type incorporates tailored designs for the constituent modules and data sources within its respective learning pipeline.
In the following sections, we present a detailed discussion of the methods under each branch of the taxonomy.

\vspace{-2mm}
\section{Points}\label{sec:points}
Points are the most fundamental geospatial objects, forming the basic component of polylines and polygons. Each point is associated with a geo-location and associated features, such as text descriptions and categories. Examples of point data include location-based point sensors, road network intersections, and points-of-interest (POIs). While extensive research has been conducted on modeling point data for forecasting tasks related to traffic, weather, and environment~\cite{unist, w-mae}, most studies adopt an end-to-end training method to improve forecasting accuracy, with no emphasis on representation learning. Recent research on self-supervised representation learning techniques for point data focuses primarily on POI data due to their rich semantic information. Consequently, this survey focuses on POI data, which plays a crucial role in understanding user mobility and urban functionality.
The surveyed studies for POI data are listed in Table~\ref{tab: poi}.

\subsection{Points of Interest}
\subsubsection{Intrinsic Attributes}
POIs refer to the semantic locations in location-based services that users might want to visit. Examples of POIs include restaurants, stores, schools, etc. Each POI is associated with geographical coordinates $l$ (i.e., geo-location) and usually a number of features $x=\{name, categories, reviews\}$, including the POI's name, one or multiple categories, and possibly a set of user reviews.

Since the geo-location and features are in different formats, existing methods~\cite{mgeo,SIGMOD23_geokg,spabert} often employ separate encoders to extract essential information from geo-location~\cite{survey_geoencoding}, categories, and other text features~\cite{BERT}. The encoded features are fused in a subsequent model, such as multi-layer perceptrons (MLP) or attention-based models, to obtain the POI representations~\cite{mgeo,SIGMOD23_geokg,spabert}. The objective of SSL for POI is to learn an encoder $f_\theta$ that can acquire a $d$-dimension latent representation of any input POI $p$, denoted by $\mathbf{e}_p = f_\theta(p) \in \mathbb{R}^d$, such that $p$'s geo-location $l$ and features $x$, as well as $p$'s context are well preserved in $\mathbf{e}_{p}$. Based on this objective, we note that scenarios of (next) POI recommendation~\cite{survey_poi-rec} are more related to specific tasks with supervised signals rather than the SSL setting. Therefore, we do not focus on these methods in this section.

\begin{table*}[tbp]
\centering
\caption{A summary of the surveyed papers on self-supervised learning for \textbf{points-of-interest}.}
{\fontsize{8}{10}\selectfont
\begin{tabular}{   
    >{\raggedright\arraybackslash}m{2.4cm}
    >{\centering\arraybackslash}m{1.2cm}
    >{\centering\arraybackslash}m{3cm}
    >{\centering\arraybackslash}m{2.7cm}
    >{\centering\arraybackslash}m{3cm}
    >{\centering\arraybackslash}m{4.1cm}}
    
    \toprule
     \textbf{Method} & \textbf{Type}  & \textbf{Data Augmentation} &   \textbf{Context Information}  & \textbf{Objective Function} & \textbf{Pretext Tasks}  \\  
    \hline

Place2vec~\cite{place2vec}    & Predictive & \tabincell{c}{Boosting the occurrence \\ of popular neighbors} & Spatial neighbors   & Skipgram  & Category prediction  \\ \hline
Semantic~\cite{semantic_poi}    & Predictive & Random walk sampling & Spatial neighbors   & Skipgram  & Category prediction  \\ \hline
MT-POI~\cite{mt_poi}    & Predictive & Random walk sampling & Spatial neighbors   & Skipgram  & Category prediction  \\ \hline
STPA~\cite{poi-annotate-ijcai23}    & Predictive & None & Spatial neighbors   & Cross-Entropy  & Category prediction  \\ \hline
DeepR~\cite{POI-competitive-kdd20}  & Predictive & None & \tabincell{c}{Spatial neighbors}    & Cross-Entropy    & POI relation prediction     \\ \hline
PRIM~\cite{POI-relation-vldb22}     & Predictive & None & Spatial neighbors   & Cross-Entropy  & POI relation prediction   \\ \hline
GeoBERT\cite{geobert}    & Predictive &     \tabincell{c}{Neighbors ordered by distance \\ Word masking}   & Spatial neighbors   &    Cross-Entropy     &     Masked word prediction    \\ \hline
SpaBERT~\cite{spabert}   & Predictive &   \tabincell{c}{Shortest paths in grid cells, \\ POI sequences ordered by \\ the distance to grid center}    & Spatial neighbors   &     Cross-Entropy   &    Masked POI prediction   \\ \hline
MGeo~\cite{mgeo}       & \tabincell{c}{Predictive, \\ Contrastive} &   \tabincell{c}{Neighbors ordered by distance \\ POI attribute masking}    & Spatial neighbors    &   \tabincell{c}{Cross-Entropy, \\ KL divergence}    &   \tabincell{c}{POI attribute prediction, \\ Distance contrast}   \\ \hline
SG-CWARP~\cite{poi-context-ijcai16}  & Predictive & None  & Check-in sequence   & \tabincell{c}{Ranking loss, \\ Skipgram}   &  POI prediction    \\\hline
CAPE~\cite{poi-hier-ijcai18}   & Predictive & None  &   Check-in sequence   &    Skipgram      &    POI ID and text prediction  \\\hline
DeepMove~\cite{deepmove}   & Predictive & \tabincell{c}{Construction of origin POI \\ pairs with the same destination}  &  Check-in sequence    &     Skipgram    &   Origin POI prediction   \\\hline
Hier-CEM~\cite{hier-embed} & Predictive &   \tabincell{c}{Hierarchical extension of \\ context categories}    & Check-in sequence  & CBOW   & Next category prediction    \\ \hline
POI2Vec~\cite{poi2vec}       & Predictive & Binary region tree construction  & Check-in sequence   & CBOW  & POI prediction \\ \hline
TALE~\cite{TALE} & Predictive & Binary temporal tree construction  & Check-in sequence   & CBOW  & POI prediction \\ \hline
LSPSL~\cite{LSPSL} & Predictive & POI attribute masking  & Check-in sequence   & Cross-Entropy  & Masked POI attribute prediction \\ \hline
MoCo-GA~\cite{MoCo-GA}  & Contrastive & Samples from previous training steps  & Co-query   & InfoNCE  & \tabincell{c}{Location-location, \\ Text-location, \\ Text-text contrast} \\ \hline
SEENet~\cite{POI-relation-kdd23} & Contrastive & Grid cell augmentation & \tabincell{c}{Temporal context, \\ Spatial neighbors, \\ Co-query, \\ check-in sequence }   & JS divergence  & \tabincell{c}{Time-aware POI relation \\ prediction} \\ \hline
CatEM~\cite{CatEM}    & Predictive & None & \tabincell{c}{Spatial neighbors, \\ Check-in sequence}  & \tabincell{c}{Skipgram}  & Category prediction  \\ \hline
DCHL~\cite{DCHL}       &  Contrastive &   Multi-view hypergraph construction    & \tabincell{c}{ Spatial neighbors \\ Check-in sequence }   &  InfoNCE    &   POI-POI contrast   \\ \hline
STKG-PLM~\cite{STKG-PLM}       & \tabincell{c}{Predictive, \\ Contrastive} &   \tabincell{c}{Spatio-temporal knowledge \\ graph construction}    & \tabincell{c}{ Spatial neighbors \\ Check-in sequence }   &  \tabincell{c}{InfoNCE \\ Cross-Entropy}   &   \tabincell{c}{POI-POI contrast \\ POI prediction}  \\ \hline
AGCL~\cite{AGCL}       & \tabincell{c}{Predictive, \\ Contrastive} &   \tabincell{c}{Spatio-temporal knowledge \\ graph, co-occurrence frequency \\ graph construction}    & \tabincell{c}{ Spatial neighbors \\ Check-in sequence }   &  \tabincell{c}{InfoNCE \\ Cross-Entropy}   &   \tabincell{c}{POI-POI constrast \\ POI prediction}  \\ \hline
POI-Enhancer~\cite{poienhancer} & Contrastive & \tabincell{c}{Geography view, \\ Sequence-time view, \\ Functional view\\ construction}  & \tabincell{c}{Spatial neighbors \\ Check-in sequence}   & InfoNCE   & POI-POI contrast \\ \hline
ERNIE-GeoL~\cite{ERNIE-GeoL} & Predictive & \tabincell{c}{Heterogeneous graph, \\ Random walk sampling, \\ Word substitution\\and masking}  & \tabincell{c}{Co-query context, \\Spatial neighbors}   & Cross-Entropy   & \tabincell{c}{Masked word prediction, \\  Geocoding prediction}  \\ 
\bottomrule
    
\end{tabular}
}
\label{tab: poi}
\end{table*}

\subsubsection{Context Information}
The intrinsic attributes only include the information of individual POIs. POIs are often located in spatial environments where diverse types of POIs are present at different distance ranges. Besides, users' check-ins or queries related to POIs indicate the dependency on the urban functions of distinct POIs. Consequently, most SSL research for POIs has been dedicated to modeling the rich context information of POIs, including spatial neighborhoods, check-in sequences, co-query context, and temporal context.

\noindent\textbf{Spatial Neighborhood}. To effectively capture the spatial context of a POI, Skip-gram~\cite{place2vec}, GNNs~\cite{poi-annotate-ijcai23,POI-competitive-kdd20,POI-relation-vldb22} and masked language modeling (MLM)~\cite{geobert,mgeo,spabert,ERNIE-GeoL} have been explored. The skip-gram methods consider the spatial neighbors as the context of a POI, analogous to the surrounding words in Word2vec \allowbreak ~\cite{word2vec}. They minimize the errors in predicting the spatial context using the target POI embeddings. For instance, Place2vec~\cite{place2vec} partitions the spatial neighborhood of a target POI into equal-distance bins and calculates a bin-wise boosting factor to increase the occurrence of popular POI categories in each bin. In this way, POI categories with a higher boosting factor will contribute more to the skip-gram learning objective. In this vein, Huang et al.~\cite{semantic_poi} propose to preserve the hierarchical structure of POI categories in learning POI category embeddings. For example, the categories Japanese Restaurants and Chinese Restaurants should be similar in the embedding space due to their resemblance of functions and the fact that they often share the same generic category, e.g., Food. They employ Laplacian eigenmaps as regularization terms to pull together the POI categories that share the same ancestor categories. This method is later extended in~\cite{mt_poi} and~\cite{CatEM} to consider categories with close semantics in different years and the check-in sequences, respectively. 

GNN-based methods construct a graph where two POIs are connected by an edge if they are close in spatial. Leveraging the ability of GNNs to acquire the local structure of a graph, these methods encode the spatial neighborhood of a POI into its latent representation. Predictive SSL methods have been proposed to predict either the intrinsic attributes of the POI or the relation between POIs. STPA~\cite{poi-annotate-ijcai23} constructs a Delaunay triangulation graph based on the distance between POIs. Each node in the graph is a POI, represented by a one-hot encoding of its category. GNN is applied to acquire a target POI's representation by aggregating the category information from its neighbors in the Delaunay triangulation graph. Subsequently, a predictive objective is employed to predict the category of the target POI, given its latent representation. For objectives that predict the relations (competitive or complementary) between POIs, the relations between POIs are constructed based on certain heuristic rules. DeepR~\cite{POI-competitive-kdd20} builds the competitive relations between two POIs if they are spatially close and frequently co-occur in the same query. DeepR employs a Heterogeneous POI Information Network (HPIN) to represent POI, brands, aspects, and their relations, and a spatial adaptive GNN to acquire the POI representations from neighbors. Given the representations of two POIs, it predicts whether they are in a competitive relation. PRIM~\cite{POI-relation-vldb22} constructs competitive relations between different categories of POIs if they frequently co-occur in the same query. Likewise, it constructs complementary relations between POIs in the same category that frequently co-occur in the same query. To predict the relations, PRIM employs the POI representations obtained through a GNN that gives importance to the neighbors based on spatial distance and feature similarity. 

Besides, numerous studies leverage the MLM idea applied in pre-trained language models to learn POI representations~\cite{geobert,spabert,mgeo}. The common idea behind MLM-based methods is to construct pseudo sentences from the spatial neighborhoods and then apply MLMs to the pseudo sentences. GeoBERT~\cite{geobert} divides the digital map into grids and proposes two methods to create a pseudo sentence from POIs residing in each grid cell. The first method creates the pseudo sentence by finding the shortest path between the two farthest POIs, passing through all POIs in the same grid cell. The second method returns the ordered sequence of POIs by their distance to the grid center. With the obtained pseudo sentences, GeoBERT treats each POI as a token and adopts the same training mechanism as existing MLMs -- predicting a masked POI in the pseudo sentence based on others. SpaBERT~\cite{spabert} creates pseudo sentences by concatenating the names of neighboring POIs. For each POI, SpaBERT finds its neighboring POIs using Geohash and sorts them by their distance to the POI. Subsequently, SpaBERT masks some or all words of a certain POI in a pseudo-sentence and predicts the masked words based on the remaining words in the sentence. In this way, the learned POI representation preserves essential information for generating the words that appear in the nearby POIs, thus capturing the underlying correlations between geo-locations and text. To better leverage the spatial context and support query-POI matching, MGeo~\cite{mgeo} designs several geospatial encoders to encode the spatial attributes of a POI, including ID, shape, map position, and relative location to neighboring spatial objects extracted from OpenStreetMap~\cite{osm}. Subsequently, MGeo employs both predictive and contrastive SSL objectives to train these encoders. The predictive SSL objective minimizes the loss of predicting the masked attribute (e.g., shape) using other attributes. The contrastive objective minimizes the difference between the actual spherical
distance and the distance computed by the representations of all pairs of POIs in the same batch. After training the encoders, MGeo fuses the text and spatial information through another MLM that predicts the masked words or spatial attributes of a POI.

\noindent\textbf{Check-in Sequence}. Location-based services allow users to share their visits at POIs. Such visits are often referred to as ``check-ins''. Each check-in record is a triplet of (\emph{u, p, t}), denoting a user $u$ visiting a POI $p$ at a certain time $t$. The check-in records for a specific user can be arranged in chronological order, forming a check-in sequence. The check-in sequences produced by users often exhibit strong dependencies between POIs. 

Most existing studies~\cite{poi-context-ijcai16,poi2vec, poi-hier-ijcai18,deepmove,hier-embed} employ skip-gram, contextual bag-of-word (CBOW), which are three typical predicative objectives, to the check-in sequences. 
Specifically, SG-CWARP~\cite{poi-context-ijcai16} learns POI representations to predict other POIs within the same check-in sequences. As a result, the co-occurren \allowbreak ces of POIs in the same check-in sequences are preserved in the POI representations. 
Building upon this, CAPE~\cite{poi-hier-ijcai18} learns POI representations that effectively predict IDs and texts of other POIs in the same sequences. In this way, the text information in the sequential context is encoded in the learned POI representations. Unlike previous methods that predict all POIs visited within a close time period, DeepMove~\cite{deepmove} focuses more on the intent of the trip and considers only the origin and destination POIs of a trip. Utilizing an origin POI, DeepMove predicts other origin POIs with the same destination POI. Consequently, POIs with similar travel intents are projected into a close region in the learned representation space. Apart from predictive methods that employ the skip-gram objective, SSL methods with the CBOW objective have been proposed to predict a check-in POI based on its context. These methods often focus on the modeling of spatio-temporal context in the check-in sequences. Hier-CEM~\cite{hier-embed} extends the check-in sequence context by associating each check-in POI with the ancestor categories. POI2Vec~\cite{poi2vec} introduces a binary region tree to enhance the modeling of check-in sequence context with spatial proximity. TALE~\cite{TALE} extends the binary region tree in POI2Vec with time information, producing a temporal tree that splits the check-ins into different time slots. In this way, TALE captures both spatial and temporal proximity in the context of a check-in.

Motivated by Masked Language Models (MLM), some recent studies propose to learn POI representation by predicting the masked check-in in a sequence. 
LSPSL~\cite{LSPSL} obtains a check-in's representation by a self-attention network, aggregating information from its attributes (e.g., location). The attribute embeddings are then pre-trained by employing each check-in's representation to predict the attributes of other check-in records in the same trajectory.

\noindent\textbf{Co-query Context}. In digital map services, users can submit text queries to search for POIs. The POIs clicked by a user in the same query are often correlated. Motivated by this, several studies have been dedicated to incorporating the co-query context in learning POI representations. ERNIE-GeoL~\cite{ERNIE-GeoL} constructs a heterogeneous graph containing query and POI nodes to capture the correlation between POIs in the same query. POIs are connected by a typed edge if they are clicked successively in the same query or co-locate in the same geographical region. A POI node is connected to its historical query nodes. On the heterogeneous graph, ERNIE-GeoL runs a random walk algorithm to create node sequences, on which it applies two predictive SSL objectives to learn the representations of queries and POIs. Specifically, the first SSL objective minimizes the prediction errors for the masked words of POI or query nodes in the node sequence. The second SSL objective minimizes the error of predicting the geocoding in the discrete global grid system~\cite{discrete_global_grid}. Besides, to enhance the learning of the text features of POIs, ERNIE-GeoL augments the text attribute of POIs by randomly swapping words with misspelled words or random words. POI relationship prediction methods, such as SEENet~\cite{POI-relation-kdd23}, use the POI queries to construct competitive and complementary relations between POIs and employ GNNs to obtain POI representations that preserve the co-query information. Unlike the previous predictive SSL methods, MoCo-GA~\cite{MoCo-GA} employs momentum contrastive learning to ensure the current POI embeddings are closer to the query embeddings than those obtained in the previous training steps.

\noindent\textbf{Temporal Context}. POIs may exhibit distinct relations at different times. For instance, a restaurant and a coffee shop might be competitive during breakfast and complimentary during  lunch time. Motivated by this, SEENet~\cite{POI-relation-kdd23} proposes to learn POI representations for different time slots. It constructs a dynamic graph of POIs with changing relations between POIs and adapts GNN to capture the intra-time and inter-time correlations between POIs. SEENet optimizes two SSL objectives simultaneously. In the first SSL objective, SEENet divides the map into grid cells and considers POI at time $t$ and its corresponding grid cell at time $t-1$ as a positive sample. In contrast, negative samples are created by randomly replacing the grid cell with a random grid cell. Subsequently, it adopts a contrastive loss~\cite{velickovic2019deep} to maximize mutual information based on the positive and negative samples. In the second SSL objective, SEENet predicts the existence of a relation between two consecutive time slots. The two SSL objective functions enable SEENet to encode time-specific features and the evolving relations between POIs in the POI representations.

\noindent\textbf{Multiple Context Sources}. Since POIs are often located within complex spatiotemporal contexts, extensive research highlights the value of incorporating multiple contextual sources when learning POI representations.
CatEM~\cite{CatEM} represents the category co-occurrence in check-in sequences by a Point-wise Mutual Information (PMI) matrix and captures the spatial neighborhood by a category proximity matrix. CatEM learns POI representations by simultaneously optimizing the matrix factorization loss on the PMI matrix and minimizing the distance between embeddings of nearby POIs. In this way, CatEM incorporates the spatial neighborhood context and the check-in sequence context. DCHL~\cite{DCHL} constructs hypergraphs from three different views, including geographical view, collaborative view, and transitional view, to incorporate the spatial neighborhood and sequential information of check-in sequences at the same time. To learn POI representations, DCHL employs a cross-view contrastive learning technique that takes the same POI from the different views as positive pairs while different POIs as negative pairs. STKG-PLM~\cite{STKG-PLM} models the POI data with a spatio-temporal knowledge graph, defining spatial neighborhoods and check-in context by pre-defined relations. It employs a knowledge graph encoder to obtain POI representations and the InfoNCE~\cite{SSL_image1} objective for minimizing the gap between POI representations checked in by the same user while maximizing those checked in by different users. Similar to STKG-PLM, AGCL~\cite{AGCL} constructs a spatio-temporal knowledge graph to capture the spatial neighborhood and transitional relations between POIs. Different from STKG-PLM, it constructs a co-occurrence frequency graph to incorporate the frequency of all users who visit one POI after another. To learn POI representations, AGCL employs both intra-graph and inter-graph contrastive learning objectives. The intra-graph contrastive objective ensures that POIs with high correlations within and across graphs are positioned closely in the embedding space. 
POI-Enhancer~\cite{poienhancer} proposes a POI embedding model to extract semantic information of POIs from large language models and employs a multi-view contrastive learning method to learn POI representations. It creates a geography view, a sequence-time view, and a functional view of POIs to incorporate the spatial, sequential, and semantic context of POIs. POIs that are located in a close region are considered positive samples in the geography view, and those that share the same category and visit patterns are considered positive samples in the functional view. In the sequence-time view, POIs visited in a close time period are regarded as positive samples. With these three views, POI-Enhancer learns a comprehensive POI representation by optimizing the InfoNCE loss.
While most studies incorporate the spatial context and the sequential context of POIs, some research has been dedicated to exploring other combinations of context sources. For instance,  
ERNIE-GeoL~\cite{ERNIE-GeoL} incorporates multiple context sources by defining a heterogeneous graph with three types of edges, i.e., Origin-to-Destination, POI co-location, and POI co-query. SEENet~\cite{POI-relation-kdd23} constructs a dynamic graph based on the spatial proximity and the temporal relation between POIs to incorporate spatiotemporal, sequential, and co-query contexts.

\vspace{-3mm}
\subsubsection{Applications} The POI representations acquired from the SSL methods can be applied to various downstream applications. Most POI representations can be used to predict specific POI attributes, such as category~\cite{poi-annotate-ijcai23} and geo-location~\cite{mgeo}. Besides, POI presentations can be directly used or used with a subsequent sequential model to predict the next POI visited by a user~\cite{poi-context-ijcai16,poi2vec, poi-hier-ijcai18}. The POI representations can also be used to answer POI queries by matching POI representations and the query representation~\cite{ERNIE-GeoL}.  
Furthermore, the representations of multiple POIs can be used for more complex geospatial data mining tasks. For instance, POI relation prediction methods~\cite{POI-competitive-kdd20,POI-relation-vldb22,POI-relation-kdd23} use the representations of two POIs to predict their competitive or complementary relations. Likewise, the POI representations are pivotal in matching POIs from different databases for entity resolution~\cite{geo-entity-resolution}. In addition, increasing studies have utilized POI embeddings in an aggregated manner for the inference of land use~\cite{semantic_poi} or land use change~\cite{mt_poi} across years. The aforementioned applications demonstrate that the SSL methods can effectively encode POIs and their contexts, thereby helping to improve the accuracy of various GeoAI tasks.

\section{Polylines}\label{sec:polylines}

Polylines serve as an important spatial data type, widely employed across diverse GeoAI applications to depict a range of features, such as contour lines, road segments, and trajectories. SSL for polylines aims to derive effective representations for such data instances, with a particular emphasis on road segments and trajectories due to their prominence in existing literature. While both road networks and trajectories can be geometrically represented as polylines, they differ fundamentally in structure and semantics. Road networks typically denote static infrastructure with topological consistency and connectivity, whereas trajectories capture dynamic movements, often annotated with timestamps and user-specific metadata. This divergence in semantics and application motivates a distinct categorization for these two data instances in our survey, enabling a more precise analysis of SSL methods tailored to each context. Consequently,
we will separately discuss these two instances of polylines, and present the specialized SSL techniques tailored to each data instance. The surveyed studies for road networks are listed in Table~\ref{tab: road networks}.

\subsection{Road Networks}
\subsubsection{Intrinsic Attributes}
Road networks are composed of connected road segments, each of which is represented as a polyline. Since the fundamental analytical units in road network studies are typically individual road segments rather than enclosed regions, road networks are classified under the category of polylines. Conceptually, road networks can be represented as graphs, allowing for the modeling and analysis of their structural and topological properties. In this regard, two primary graph-based perspectives are adopted in the research. In the first perspective, road segments themselves are treated as nodes, and the connections between these segments are treated as graph edges. In the alternative perspective, road intersections serve as nodes, while the road segments linking these intersections are treated as edges.

The graph-based formulation from these two perspectives naturally represents the topological structure of road networks. Additionally, various attributes on road networks are integrated as node or edge features within the formulation, such as geo-locations, road attributes, and intersection characteristics. As a result, existing studies on SSL for road networks generally follow the principles of SSL for graph  data. The objective is to learn geospatial encoder $f_\theta$ to obtain node representations $\{\mathbf{e}_{v}\}_{v\in \mathcal{V}} \in \mathbb{R}^{d}$ (i.e., $\{\mathbf{e}_{v}\}_{v\in \mathcal{V}}=f_\theta(\mathcal{G}$), where $\mathcal{G}$ and $\mathcal{V}$ denote the road network graph and its nodes, by employing either graph perspective in a self-supervised manner.

\begin{table*}[hbp]
\centering
\caption{A summary of the surveyed papers on self-supervised learning for \textbf{road networks}.}
{\fontsize{8}{10}\selectfont
\begin{tabular}{   
    >{\raggedright\arraybackslash}m{2.4cm}
    >{\centering\arraybackslash}m{1.2cm}
    >{\centering\arraybackslash}m{3cm}
    >{\centering\arraybackslash}m{2.7cm}
    >{\centering\arraybackslash}m{3cm}
    >{\centering\arraybackslash}m{4.1cm}}

    \toprule
     \textbf{Method} & \textbf{Type}  & \textbf{Data Augmentation} &   \textbf{Context Information}  & \textbf{Objective Function} & \textbf{Pretext Tasks}  \\  
    \hline

R2vec\cite{road_BigData18}    & Predictive                        & Random walk sampling & None & Skipgram & Context neighbor prediction              \\  \hline
    IRN2Vec~\cite{road_IRN2Vec} & Predictive & Random walk sampling & Road attributes & Cross-Entropy & Road attribute prediction \\ \hline
    RFN~\cite{road_RFN} & Predictive & None & Road attributes & Cross-Entropy & Graph reconstruction \\ \hline
    RN2Vec~\cite{road_RN2Vec} & Predictive & Random walk sampling & \tabincell{c}{Road attributes\\ Spatial features}  & Skipgram & Road attribute prediction \\ \hline
    HyperRoad~\cite{road_HyperRoad} & Predictive & Hypergraph construction & Road attributes & \tabincell{c}{Cross-Entropy, \\ Skipgram} & \tabincell{c}{Graph\&Hypergraph reconstruction, \\ Road attribute prediction, \\ Hyperedge Classification} \\ \hline
    SARN~\cite{road_SARN} & Contrastive & Edge perturbation & \tabincell{c}{Spatial features, \\ Road attributes} & InfoNCE & Road-road, Road-region contrast \\ \hline
    HNRN~\cite{road_HNRN} & Predictive & None & Trajectory & Cross-Entropy & Graph reconstruction \\ \hline
    Toast~\cite{road_Toast} & Predictive & \tabincell{c}{Road segment masking,\\ Random walk sampling, \\ Similar trajectory generation}  & Trajectory & \tabincell{c}{Cross-Entropy, \\ Skipgram} & \tabincell{c}{Context neighbor\&attribute prediction, \\ Masked road recovery, \\ Trajectory discrimination} \\ \hline
    DyToast~\cite{Dytoast} & Predictive & \tabincell{c}{Road segment masking,\\ Random walk sampling, \\Similar trajectory generation}  & Trajectory & \tabincell{c}{Cross-Entropy \\ Skipgram} & \tabincell{c}{Time-aware context neighbor\\ \&attribute prediction, \\ Masked road recovery, \\ Trajectory discrimination} \\ \hline
    TRACK~\cite{road_track} & \tabincell{c}{Predictive,\\ Contrastive} & \tabincell{c}{Road segment masking,\\ Traffic state masking, \\Similar trajectory generation}  & Trajectory & \tabincell{c}{Cross-Entropy, MAE, \\InfoNCE} & \tabincell{c}{Time-aware masked road\& \\traffic state recovery, \\ Traffic state-road matching, \\ Trajectory discrimination} \\ \hline
    TrajRNE~\cite{road_TrajRNE} & Predictive & Transition view construction & Trajectory & Cross-Entropy, MAE & \tabincell{c}{Graph reconstruction, \\ Road attribute prediction} \\ \hline
    JCLRNT~\cite{road_JCLRNT} & Contrastive & \tabincell{c}{Transition view construction,\\ Detour generation} & Trajectory & JS divergence & \tabincell{c}{Road-road, Road-trajectory, \\ Trajectory-trajectory contrast} \\ \hline
    USPM~\cite{road_KDD24} & \tabincell{c}{Predictive,\\Contrastive} & None & SVI & \tabincell{c}{InfoNCE, \\ Cross-Entropy} & \tabincell{c}{Road-image contrast, \\ Road attribute prediction} \\ \hline
    Garner~\cite{road_Garner} & Contrastive & Multi-view graph construction & \tabincell{c}{SVI} & JS divergence & Road-road contrast \\ \bottomrule
    
\end{tabular}
}

\label{tab: road networks}
\end{table*}

Early approaches generally adopt predictive SSL with various training objectives. Initial exploration into this field~\cite{road_BigData18} applies node2vec~\cite{node2vec}, a self-supervised graph representation learning method based on skip-gram training, directly to road networks. It predicts road segments within a context window derived from random walks sampled on road networks~\cite{word2vec}. The derived representations are demonstrated to be effective on several road classification tasks. Building upon this groundwork, IRN2Vec~\cite{road_IRN2Vec} targets at learning representations for intersections by treating intersections as graph nodes. This method refines node2vec by employing shortest-path random walk sampling and selecting positive intersection pairs based on defined road path distances for skip-gram training. Moreover, it leverages intersection-specific attributes (e.g., intersection tags and types), to enrich the selection of positive road segment pairs. Another research line explores GNN with reconstruction-based objectives. For example, RFN~\cite{road_RFN_short, road_RFN} applies GNN to both perspectives of the road network graphs, fusing features from both road segments and intersections, including zone categories, road types and intersection angles, to derive representations capable of reconstructing the original graphs. HyperRoad~\cite{road_HyperRoad} expands upon the vanilla graph structure by constructing a corresponding hypergraph, where hyperedges represent the road segments within the polygons produced through a map segmentation algorithm~\cite{mapsegmentation}. GNN is further enhanced by a dual-channel attention mechanism that operates on both the original graph and the hypergraph. This method is trained to not only construct both the original graph and its corresponding hypergraph, but also to perform a hyperedge classification task.

\subsubsection{Context Information}
Apart from the topological structure and the attributes associated with road segments and intersections, road networks contain rich context information that can be leveraged to enhance the extraction of semantic knowledge. We introduce several types of context information utilized in existing methods, where road segments are mainly regarded as graph nodes.

\paratitle{Spatial Attributes}. Different from conventional graphs that primarily focus on connectivity, road networks are inherently defined within a geospatial context, exhibiting spatial features such as coordinates, lengths, and angles which are beneficial in modeling efforts~\cite{road_RFN, road_HyperRoad}. Accordingly, spatial information has been effectively utilized as context knowledge in various methods. RN2Vec~\cite{road_RN2Vec} extends IRN2Vec to obtain representations for both intersections and road segments by selecting spatially nearby positive pairs among roads and intersections in skip-gram training.   
SARN~\cite{road_SARN} constructs a weighted adjacency matrix that reflects the spatial proximity among road segments. The matrix is inferred from both road connectivity and the spatial and angular distances between road segments. Utilzing this matrix, SARN employs a GNN-based contrastive learning approach~\cite{graphCL} with graph augmentation techniques, deriving similar representations for the identical road segments from two distinct augmented graphs.

\paratitle{Trajectories}. Trajectories traveled on road networks act as valuable data sources that provide travel-related semantics beyond the topological structure. 
HNRN~\cite{road_HNRN} develops a hierarchical GNN framework to model relationships from individual road segments to structural regions and further extending to functional zones. It involves incorporating trajectories to derive connectivity matrix in structural regions, and aims to reconstruct the connectivity matrix for the base-level road segments derived from the higher level of region and functional representations. 
Toast~\cite{road_Toast} is the initial effort to explicitly utilize detailed trajectories to enhance road network representation learning through predictive SSL. It equips skip-gram training with additional traffic and attribute prediction for context neighbors. Besides, it proposes two novel trajectory pre-training tasks within BERT framework~\cite{BERT}: route recovery, which recovers a sequence of masked road segments, and trajectory discrimination, which assesses whether a trajectory is authentic or simulated through random walk sampling on the road network graph. These tasks enable the encoding of transition patterns and long-term dependencies intrinsic in road networks. DyToast~\cite{Dytoast} further extends this method by incorporating temporal consideration into the representations. It modifies the skip-gram training and the BERT framework by integrating parameterized trigonometric functions to capture dynamics and evolution in time dimension. 
Similarly, TRACK~\cite{road_track} considers temporal dynamics for road segment representations by integrating trajectory data-based transition probabilities into the GAT attention mechanism while modeling traffic states for road networks. Building on Toast pretext tasks, TRACK further employs masked traffic state imputation and prediction tasks for training dynamic road segment representations. Besides, it further incorporates a co-attentional transformer encoder with a gravity-based attention mechanism and a trajectory-traffic state matching task, ensuring mutual reinforcement between these two data sources. 
JCLRNT~\cite{road_JCLRNT} proposes three types of objectives based on contrastive SSL: road-road, trajectory-trajectory and road-trajectory contrastive loss. Each type is designed to differentiate entity pairs that share relatedness (e.g., road segments frequently traveled within trajectories) with those that do not. TrajRNE~\cite{road_TrajRNE} utilizes the road transition matrix derived from trajectories in GNN aggregation function, and employs the objective of reconstructing the original topological structure of road networks from this transition matrix.  

\paratitle{Street View Images}. Street view images (SVIs), which are available from various map services, provide high-resolution visual perspectives of road networks. These images capture the surroundings and configurations of different road segments, inherently encoding rich urban semantics and insights. USPM~\cite{road_KDD24} utilizes pre-trained image encoders to extract the representations for these images and derives a road segment representation aggregated from all associated images. The road segment representation and each of its associated images are treated as a positive pair for contrastive learning. Moreover, USPM further enhances the representations by incorporating  textual descriptions of the images and applies GNN on the topological graph of road networks. The final representations are trained with the objective of road attribute prediction. Garner~\cite{road_Garner} extends beyond the idea that nearby road segments should exhibit similar representations. It seeks to also derive similar representations for road segments that display similar geographical configurations as evidenced by street view images. To achieve these two goals, after extracting and aggregating representations from image encoders for each road segment, Garner employs a dual contrastive learning objective, in which GNNs are applied to distinguish a road segment representation within three graphs, where edges represent the topological structure, nearest neighbors and similar configurations respectively.

\subsubsection{Applications} The representations of road networks, derived from either predictive or contrastive SSL objectives, provide task-agnostic inputs that can be directly utilized or fine-tuned with geospatial encoder, for downstream applications. For example, they are directly applicable in classifying attributes of road segments or intersections with simple models (e.g., MLP), such as road type~\cite{road_TrajRNE}, lane number~\cite{road_HyperRoad}, and intersection tags~\cite{road_IRN2Vec, road_RN2Vec}. Besides, these representations are similarly utilized to infer the traffic status of road segments, including metrics like average speed~\cite{road_Toast,road_JCLRNT} and speed limits~\cite{road_RFN}. They also support the efficient vector computation for road network-based queries, such as calculating shortest path distances~\cite{road_SARN}. Furthermore, road segment representations serve as the effective start point for training models that involve map-matched trajectories, for applications such as destination prediction~\cite{road_HNRN} and anomalous sub-trajectory detection~\cite{subtraj}. By enabling diverse analytical tasks related to road networks, these representations offer substantial potential to improve understanding and decision-making within the domain of transportation infrastructure.

\subsection{Trajectory}
\subsubsection{Intrinsic Attributes} A trajectory, composed of a sequence of sampled points that represent the path of a moving object, can essentially be conceptualized as a polyline. Based on the definition, the context feature associated with each point in trajectory data instance includes timestamps, along with potentially additional textual 

\onecolumn
\begin{center} 
{\fontsize{8}{10}\selectfont
\begin{longtable}{   
    >{\raggedright\arraybackslash}m{2.4cm}
    >{\centering\arraybackslash}m{1.2cm}
    >{\centering\arraybackslash}m{3cm}
    >{\centering\arraybackslash}m{2.7cm}
    >{\centering\arraybackslash}m{3cm}
    >{\centering\arraybackslash}m{4.1cm}}
     \caption{A summary of the surveyed papers on self-supervised learning for \textbf{trajectory}.} \label{tab: trajectory}\\

    \toprule
     \textbf{Method} & \textbf{Type}  & \textbf{Data Augmentation} &   \textbf{Context Information}  & \textbf{Objective Function} & \textbf{Pretext Tasks}  \\  
    \hline
    \endfirsthead

\multicolumn{6}{c}{{\bfseries Table \thetable\ A summary of the surveyed papers on self-supervised learning for \textbf{trajectory} (continued).}} \\
    \toprule
     \textbf{Method} & \textbf{Type}  & \textbf{Data Augmentation} &   \textbf{Context Information}  & \textbf{Objective Function} & \textbf{Pretext Tasks}  \\  
    \hline
    \endhead

    \midrule
    \multicolumn{6}{r}{{Continued on next page}} \\ 
    \endfoot

    \bottomrule
    \endlastfoot

   TCDRL\cite{traj_IJCNN17}    & Predictive                        & \tabincell{c}{Sliding window \\ feature extraction} & Grid partition & Cross-Entropy & Trajectory reconstruction              \\  \hline
    t2vec~\cite{traj_t2vec} & Predictive & \tabincell{c}{Point dropping, \\ Spatial distortion} & Grid partition & Spatial-aware Cross-Entropy & Trajectory reconstruction \\ \hline
    GM-VSAE~\cite{traj_ICDE20} & Predictive & None & Grid partition & Cross-Entropy & Variational trajectory reconstruction \\ \hline
    Geo-Tokenizer~\cite{traj_ECML23} & Predictive & None & Multi-scale grid partition & Multi-scale Cross-Entropy & \tabincell{c}{Autoregressive next grid \\prediction} \\ \hline
    AdvTraj2vec~\cite{traj_CIKM22_adv} & Predictive & \tabincell{c}{Point dropping,\\ Embedding perturbation} & Grid partition & \tabincell{c}{Spatial-aware \\ Cross-Entropy, \\Adversarial loss} & \tabincell{c}{Trajectory reconstruction, \\ Adversarial learning} \\ \hline
    E$^{2}$DTC~\cite{traj_ICDE21} & Predictive & \tabincell{c}{Point dropping, \\ Spatial distortion} & Grid partition &\tabincell{c}{Spatial-aware \\ Cross-Entropy, \\ Triplet loss, KL divergence} & \tabincell{c}{Trajectory reconstruction, \\Triplet margin similarity, \\Self-clustering} \\ \hline
    STPT~\cite{traj_ICDM23} & Predictive & None & \tabincell{c}{Grid partition}  & Cross-Entropy & \tabincell{c}{Sub-trajectory discrimination} \\ \hline
    ST-t2vec~\cite{traj_TIST21} & Predictive & \tabincell{c}{Point dropping, \\ Spatial\&Temporal distortion}  & \tabincell{c}{3D temporal \\ grid partition} & \tabincell{c}{MSE} & \tabincell{c}{Trajectory reconstruction, \\ Pairwise\&Pattern similarity} \\ \hline
    RSTS~\cite{traj_WWWJ23} & \tabincell{c}{Predictive, \\ Contrastive} & \tabincell{c}{Point dropping,\\ Spatial\&Temporal distortion} & \tabincell{c}{3D temporal \\ grid partition}  & \tabincell{c}{Spatial-aware \\ Cross-Entropy,\\Triplet loss} & \tabincell{c}{Trajectory reconstruction, \\Triplet margin similarity} \\ \hline
    DeepTEA~\cite{traj_VLDB22} & Predictive & None & \tabincell{c}{Grid partition,\\ Historical traffic} & Cross-Entropy & \tabincell{c}{Variational trajectory \\ reconstruction} \\ \hline
    UniTraj~\cite{unitraj} & Predictive & \tabincell{c}{Point dropping} & \tabincell{c}{Point coordinates}  & Cross-Entropy & \tabincell{c}{Trajectory reconstruction} \\ \hline
    CL-Tsim~\cite{traj_CIKM22_contrastive} & Contrastive & \tabincell{c}{Point dropping, \\ Spatial distortion} & Grid partition & InfoNCE & \tabincell{c}{Trajectory-trajectory contrast} \\ \hline
    TrajCL~\cite{traj_ICDE23_contrastive} & Contrastive & \tabincell{c}{Point dropping, \\Spatial distortion, \\Trajectory trimming \\ \& simplification} & \tabincell{c}{Grid partition,\\Point coordinates}  & InforNCE & \tabincell{c}{Trajectory-trajectory contrast} \\ \hline
    KGTS~\cite{traj_AAAI24_contrastive} & Contrastive & \tabincell{c}{Grid deletion, \\Grid alternation} & \tabincell{c}{Grid partition}  & InfoNCE & \tabincell{c}{Trajectory-trajectory contrast} \\ \hline
    CSSRNN~\cite{traj_IJCAI17_road} & Predictive & Map matching & Road networks  & Cross-Entropy & Autoregressive road prediction \\ \hline
    Traj2Vec~\cite{traj_TIST_road} & Predictive & Map matching & Road networks & Cross-Entropy, MSE & \tabincell{c}{Trajectory reconstruction,\\Travel time estimation}\\ \hline
    PT2Vec~\cite{traj_DASFAA23_road} & Predictive & \tabincell{c}{Map matching, \\ Road network partition} & \tabincell{c}{Road networks}  & Cross-Entropy & \tabincell{c}{Trajectory reconstruction} \\ \hline
    JGRM~\cite{traj_WWW24_road} & Predictive & \tabincell{c}{Map matching, \\ Road segment masking} & \tabincell{c}{Road networks}  & Cross-Entropy & \tabincell{c}{Masked road recovery, \\Trajectory discrimination} \\ \hline
    ST2Vec~\cite{traj_KDD22_road} & Contrastive & Map matching & Road networks & Triplet loss & \tabincell{c}{Triplet margin similarity} \\ \hline
    GRLSTM~\cite{traj_AAAI23_road} & Contrastive & \tabincell{c}{Map matching} & \tabincell{c}{Road networks}  & Triplet loss & \tabincell{c}{Triplet margin similarity} \\ \hline
    PIM~\cite{traj_IJCAI21_road} & Contrastive & Map matching & Road networks & \tabincell{c}{JS divergence} & \tabincell{c}{Road-road, Road-trajectory \\contrast} \\ \hline
    MMTEC~\cite{traj_TKDE_road} & Contrastive & \tabincell{c}{Map matching, \\ Continuous Trajectory} & \tabincell{c}{Road networks}  & Maximize entropy encoding & \tabincell{c}{Trajectory-trajectory contrast} \\ \hline
    HMTRL~\cite{traj_VLDBJ_road} & \tabincell{c}{Predictive, \\Contrastive} & Map matching & Road networks & Cross-Entropy, MSE & \tabincell{c}{Masked attribute prediction, \\ Trajectory-trajectory contrast} \\ \hline
    START~\cite{traj_ICDE23_road} & \tabincell{c}{Predictive,\\Contrastive} & \tabincell{c}{Map matching, Dropout, \\ Temporal distortion, \\ Trajectory trimming, \\ Road segment masking} & Road networks & \tabincell{c}{Cross-Entropy, \\ InfoNCE} & \tabincell{c}{Masked road recovery, \\ Trajectory-trajectory contrast} \\ \hline
    LightPath~\cite{traj_KDD23_road} & \tabincell{c}{Predictive, \\Contrastive} & \tabincell{c}{Map matching, \\ Road segment masking} & \tabincell{c}{Road networks}  & Cross-Entropy & \tabincell{c}{Masked road recovery, \\ Multi-view trajectory matching} \\ \hline
    GREEN~\cite{traj_KDD25_road} & \tabincell{c}{Predictive, \\Contrastive} & \tabincell{c}{Map matching, \\ Road segment masking} & \tabincell{c}{Grid partition, \\ Road networks}  & \tabincell{c}{Cross-Entropy, \\InfoNCE} & \tabincell{c}{Masked road recovery, \\ Road-grid view trajectory contrast} \\ \hline
    CTLE~\cite{traj_AAAI21_semantic} & Predictive & \tabincell{c}{Point masking,\\Time masking} & \tabincell{c}{Location semantics}  & Cross-Entropy & \tabincell{c}{Masked location\&time recovery} \\ \hline
    At2vec~\cite{traj_TKDE23_semantic} & Predictive & \tabincell{c}{Point dropping} & \tabincell{c}{Location semantics}  & \tabincell{c}{Spatial-temporal \\-activity-aware \\ Cross-Entropy} & Trajectory reconstruction \\ \hline
    SML-LP~\cite{traj_KBS_semantic} & Contrastive & \tabincell{c}{Point dropping,\\Location alternation} & \tabincell{c}{Location semantics}  & InfoNCE & \tabincell{c}{Location embedding contrast} \\ \hline
    CLMTR~\cite{traj_clmtr} & Contrastive & \tabincell{c}{Point dropping, \\Spatial distortion, \\Trajectory trimming \\\& simplification} & \tabincell{c}{Location semantics}  & \tabincell{c}{InfoNCE} & \tabincell{c}{Spatial-temporal\\-textual feature contrast,\\ Trajectory-trajectory contrast} \\ \hline
    CACSR~\cite{traj_AAAI23_semantic} & Contrastive & \tabincell{c}{Location embedding \& latent \\ space perturbation} & \tabincell{c}{Location semantics}  & \tabincell{c}{InfoNCE} & Latent representation contrast \\ \hline
    MVTraj~\cite{traj_mvtraj} & \tabincell{c}{Predictive,\\ Contrastive} & \tabincell{c}{Map matching, \\ Road segment masking,\\ Point dropping} & \tabincell{c}{Grid partition,\\ Road networks,\\ Location semantics}  & \tabincell{c}{Cross-Entropy, \\ InfoNCE} & \tabincell{c}{Masked grid/road recovery, \\ Multi-view trajectory \\cross contrast} \\ \hline
    PTrajM~\cite{traj_ptrajm} & \tabincell{c}{Contrastive} & \tabincell{c}{Map matching,\\ Point allocation} & \tabincell{c}{Road networks,\\ Location semantics}  & \tabincell{c}{InfoNCE} & \tabincell{c}{Trajectory-poi view contrast, \\ Trajectory-road view contrast} \\ \hline
    RED~\cite{traj_red} & \tabincell{c}{Predictive} & \tabincell{c}{Map matching} & \tabincell{c}{Road networks,\\ User information}  & \tabincell{c}{Cross-Entropy} & \tabincell{c}{Masked road recovery, \\ Next segment prediction} \\ \hline
    MM-Path~\cite{traj_KDD25_mmpath} & \tabincell{c}{Predictive, \\Contrastive} & \tabincell{c}{Map matching,\\ Road segment masking} & \tabincell{c}{Road networks,\\ Road images}  & \tabincell{c}{Triple loss, \\InfoNCE, \\ Cross-Entropy} & \tabincell{c}{Masked road recovery,\\ Multi-granularity road-image \\ view contrast, \\ Fused road-image view contrast} 
    
\end{longtable}
}\end{center}
\twocolumn

\noindent content, and semantic tags, etc. 
The objective of SSL for trajectory data is to learn a representation produced by geospatial encoder $f_\theta$ for any given trajetocry $\mathcal{T}$:  $\mathbf{e}_{T}= f_\theta(\mathcal{T}) \in \mathbb{R}^{d}$. The sequential nature of a trajectory, marked by its spatio-temporal point sequence, represents its most fundamental intrinsic attributes. Accordingly, sequential models are used to model trajectory data, effectively capturing the transition patterns and long-term dependencies inherent in this data instance. The surveyed studies for trajectory are listed in Table~\ref{tab: trajectory}.

For predictive SSL methods, trajectories are typically trained to reconstruct their original sequence from a corrupted version of the input. For example, TCDRL~\cite{traj_IJCNN17} proposes to extract local features within sliding windows applied on trajectory records, such as speeds and rate of turns. These features are then processed using a sequence-to-sequence encoder-decoder architecture~\cite{seq2seq} based on RNN to reconstruct the local features for each window.  t2vec~\cite{traj_t2vec} refines this reconstruction process by aligning with the tokenization paradigm in natural language. It partitions the geospatial space into regular and square-sized grids and match the coordinates to their corresponding grids (tokens). Moreover, t2vec introduces downsampling and point distortion in the input sequence, and aims to reconstruct the original trajectory with spatial proximity aware loss that penalizes more for the predicted tokens that deviate significantly from the correct grids. In such an encoder-decoder framework, the vector produced by encoder part can be treated as the trajectory representation. 

This framework is enhanced by integrating a variety of modifications in terms of model architecture and loss functions, such as integrating via variatioanl inference~\cite{traj_ICDE20} or self-attention \allowbreak ~\cite{traj_ICDE21_selfattn}.
Moreover, Geo-Tokenizer~\cite{traj_ECML23} reduces the number of grids to be trained by representing a location as a combination of multiple shared grids at several granular scales. It  utilizes the objective of predicting the grids for the next token at each scale in a hierarchical way. Furthermore, AdvTraj2vec~\cite{traj_CIKM22_adv} aims to learn more robust trajectory representations through adversarial training. It adds perturbations to the token embeddings of the input sequence, with the magnitude of these perturbations guided by generative adversarial network~\cite{GAN} to ensure the effects are neither too large or too small. E$^{2}$DTC incorporates self-training via soft cluster assignments~\cite{traj_ICDE21} as an auxiliary loss into the reconstruction process. STPT~\cite{traj_ICDM23} performs a sub-trajectory discrimination loss that differentiates whether pairs of sub-trajectory representations originate from the same source. 

In addition to considering spatial dimension associated with trajectories, several SSL methods leverage temporal dimension into the reconstruction process. The method in~\cite{traj_TIST21} expands the square-sized spatial partitions to 3D spatio-temporal grids for each point. It accordingly adapts the reconstruction loss to account for temporal effects, imposing heavier penalties for reconstructed results with larger temporal discrepancies. Similarly, RSTS~\cite{traj_WWWJ23} also employs 3D spatio-temporal grids and applies a linear combination of spatial and temporal distances for reconstructed results as the final loss. Besides, DeepTEA~\cite{traj_VLDB22} utilizes Convolutional-LSTM~\cite{ConvLSTM} to model historical traffic conditions reflected in holistic trajectories, thus enhancing the variational inference by providing additional hints into dynamic patterns for each trajectory.  Recently, UniTraj~\cite{unitraj} introduces a large-scale, global trajectory dataset to train a universal trajectory model capable of generalizing across diverse tasks and regions, thus utilizing pure spatial coordinates without partitioning. To manage variations in sampling rates, it employs dynamic resampling and utilizes multiple masking strategies, including random, block, key point, and last-N masking, for data augmentation. The model adopts an encoder-decoder architecture with Rotary Position Embedding \cite{rope} to model trajectory sequences, and aims to predict masked segments for training.

In the category of contrastive SSL methods, standard mutual information maximization objectives are applied to produce similar/dissimilar representations for views derived from the same/different trajectories with various data augmentation operations. CL-Tsim~\cite{traj_CIKM22_contrastive} primarily uses point distortion to generate positive pairs of trajectories for contrastive learning. Expanding on this, TrajCL~\cite{traj_ICDE23_contrastive} introduces three additional operations, namely point masking, trajectory truncating and trajectory simplification, to enhance the diversity of the patterns for positive pairs. Besides, it proposes a dual-feature self-attention-based encoder to process not only the grid sequence, but also the spatial attributes for points, such as coordinates, angles and lengths. KGTS~\cite{traj_AAAI24_contrastive} further implements GNN to consider the interactions of neighboring grids, and includes grid deletion and movement operations to both the entire trajectories and partial trajectories to create positive pairs.

\subsubsection{Context Information}
In addition to the intrinsic spatial and temporal attributes, we discuss two distinct scenarios where trajectories are modeled under specific constraints and characteristics with further context information.

\paratitle{Road Networks}. While trajectories offer travel-related semantics for road networks, road networks in turn serve as complementary elements that impose latent topological constraints for trajectories. Specifically, when path information alongside road networks is emphasized, trajectories are usually map-matched to road networks~\cite{mapmatching1, mapmatching2}. This process transforms the trajectory from a sequence of points to a sequence of road segments for subsequent modeling. 

Several predictive SSL methods for map-matched trajectories utilize objectives similar to those used in trajectories with grid partitions, such as the reconstruction of road segments~\cite{traj_TIST_road,traj_DASFAA23_road} and next road prediction~\cite{traj_IJCAI17_road}. Besides, tailored objectives are devised to effectively incorporate the knowledge within road networks. For example,  JGRM~\cite{traj_WWW24_road} aims to recover the road segments masked from the complete path. It also considers the original point sequence and adopts another objective of differentiating whether pairs of representations are derived from the same trajectory with both the road segment sequence and the point sequence.

For contrastive SSL methods applied to map-matched trajectories, ST2Vec~\cite{traj_KDD22_road} adopts co-attention mechanism to merge spatial and temporal embeddings, followed by LSTM sequence modeling. This model employs energy-based margin functions (i.e., triplet loss) to enforce higher similarities for positive pairs which follow similar routes. GRLSTM~\cite{traj_AAAI23_road} combines GNN and graph embedding techniques~\cite{TransH} to handle sequence inputs and augments LSTM with residual connections to generate trajectory representations. It employs similar triplet loss at both the point level and the trajectory level. In addition, PIM~\cite{traj_IJCAI21_road} treats trajectories that share the same source and destination as positive pairs, and aims to maximize mutual information over these pairs. MMTEC~\cite{traj_TKDE_road} utilizes both a discrete sequence model based on Transformer, and a continuous model formulated as neural controlled differential equation to generate representations from two views. It applies a  contrastive learning objective of maximizing entropy coding between two views.  

Moreover, both contrastive and predictive SSL can be simultaneously employed with a single framework. HMTRL~\cite{traj_VLDBJ_road} employs the strategies to predict road attributes and traverse time for road segments, as well as employing full trajectory and its sub-trajectory for contrastive learning. START~\cite{traj_ICDE23_road} incorporates the trajectory pattern in GNN aggregation and enhances the Transformer with a time-sensitive self-attention mechanism. Furthermore, it utilizes masked road segment recovery as the predictive task, and enhances its contrastive learning aspect by employing four distinct data augmentation operation -- trajectory trimming, masking, temporal shifting, and dropout-- to generate positive pairs. Similarly, LightPath~\cite{traj_KDD23_road} utilizes masked road path recovery as its predictive task. For its contrastive aspect, LightPath achieves pairwise matching through the use of dual encoders and varied dropping ratios to generate positive pairs from the same trajectories.  GREEN~\cite{traj_KDD25_road} integrates a grid encoder based on CNN and a road encoder based on GNN to encode trajectories from two perspectives. It employs contrastive training to align the representations from both encoders for the same trajectory while using masked road segment recovery as a predictive task.

\paratitle{Semantic Information}. Trajectories may carry rich semantic meaning when composed of check-in sequence at specific POIs, which provide concrete location names and related activities (i.e., categories) but result in sparser sequences. Several SSL methods are developed to tackle semantic trajectories. CTLE~\cite{traj_AAAI21_semantic} designs two predictive tasks, namely masked location recovery and masked hour prediction, with a sinusoidal temporal encoding technique incorporated into Transformer to derive representations. At2vec~\cite{traj_ICWS19_semantic, traj_TKDE23_semantic} utilizes an encoder-decoder framework for sequence reconstruction, and adopts multi-level attention to consider the importance of semantic information at different locations. SML-LP~\cite{traj_KBS_semantic} applies trajectory augmentation techniques that modify several locations within close spatial and temporal proximity to form positive pairs, and aims to maximize the mutual information between the LSTM-derived hidden representations and the future location representations. CLMTR~\cite{traj_clmtr} encodes trajectory by leveraging textual descriptions of POIs processed with BERT alongside spatial and temporal embeddings to capture geographical proximity and periodicity. It employs contrastive learning at two levels: intra-trajectory, where textual features are contrasted with fused spatiotemporal features within the same trajectory, and inter-trajectory, where nearest-neighbor trajectories serve as positive pairs while distant ones act as negatives.  Furthermore, CACSR~\cite{traj_AAAI23_semantic} innovates by generating challenging positive and negative pairs in the representation space, rather than input sequence, via adversarial perturbations. We note that while SSL techniques are integrated into the modeling of semantic trajectories in other studies~\cite{traj_CIKM23_supervised, traj_TSC_supervised}, the main target and its objective derives from supervised applications. Therefore, these specific studies are not further discussed in our context.  

\paratitle{Multiple Context Sources}. The extensive studies have revealed that different types of context information provide distinct and complementary insights for trajectory data representation. For instance, spatial coordinates and grid partitions offer high-resolution positional accuracy, while road networks impose movement constraints and reveal structural mobility patterns. Additionally, location semantics provide valuable information regarding regional functionalities and human activity patterns. To this end, recent research has increasingly focused on integrating multiple context sources simultaneously, recognizing that their combination enables a more comprehensive and multi-dimensional understanding of trajectory data. MVTraj~\cite{traj_mvtraj} integrates context information from multiple views, including GPS trajectories, road networks and POIs. It employs three encoders to capture point sequences, road network paths, and grid-based semantic features, utilizing a hierarchical multi-modal interaction with twelve attention streams for cross-view fusion. The model incorporates masked part recovery as a predictive task and applies contrastive learning to align trajectory representations across different views. PTrajM~\cite{traj_ptrajm} utilizes the Mamba state-space model~\cite{mamba} to capture continuous movement behaviors and infer travel purposes from road network and POI semantic views by assigning data points to POIs. It extracts road-based and POI-based trajectory representations using pre-trained textual embeddings and Transformer-based encoders. A contrastive learning approach is then employed to align trajectory representations extracted from the state-space model with their corresponding road network and POI-based representations. Several studies have expanded trajectory representation learning by integrating additional contextual information. RED~\cite{traj_red} incorporates user information alongside spatial and temporal encodings to enhance trajectory representation within a road network context. It employs dual Transformer models for next-segment prediction and masked road segment prediction, adapting the latter to selectively mask non-crucial path segments as a pretext task. MM-Path~\cite{traj_KDD25_mmpath} combines road network-based trajectories with corresponding road segment images across three granularities: road intersection level, road sub-path level, and entire path level. It employs Transformer-based encoders for both modalities and introduces a GNN-based cross-modal fusion to integrate image and road network representations. The model utilizes masked road segment recovery as a predictive task and applies contrastive learning by aligning representations across modalities and granularities.

\subsubsection{Applications}
Trajectory representations can be directly utilized or fine-tuned for downstream applications. These representations, regardless of their specific context information, facilitate a variety of operations due to their vectorized format, including similarity computation~\cite{traj_t2vec, traj_VLDB24} and clustering~\cite{traj_ICDE21, traj_IJCNN17}. Moreover, they can be utilized to support diverse applications, such as transportation mode prediction~\cite{traj_ECML23}, driver status inference~\cite{traj_ICDM23}, and anomalous trajectory detection~\cite{traj_ICDE20}. In scenarios involving map-matched trajectories, these representations prove particularly valuable in applications focused on path analysis, such as travel time estimation~\cite{traj_ICDE23_road}, path ranking~\cite{traj_KDD23_road}, and destination prediction~\cite{traj_TKDE_road}. Semantic trajectory representations trained with SSL methods are typically fine-tuned to enhance performance in next location prediction~\cite{traj_AAAI21_semantic}, and can be uniquely applied in trajectory user-linking task~\cite{traj_AAAI23_semantic, traj_KBS_semantic}. These applications demonstrate the broad utility and adaptability of SSL for trajectories in advancing the understanding and analysis of mobility patterns and intelligent transportation systems.
\section{Polygons}\label{sec:polygons}

Polygonal representations offer detailed 2D shapes of geospatial objects. In principle,  many geospatial objects like POIs and road segments could be represented this way, while the effort to collect precise polygonal data limits its use. In this survey, we focus on urban regions, i.e., small urban areas which serve as the analytical units for a wide array of urban analytical tasks, e.g., for land use, population density, and house price.

Urban regions are partitioned, e.g., by road networks, grids, or administrative boundaries like Singapore Subzones~\cite{sg_subzone} and NYC Census Tracts~\cite{nyc_subzone}. These regions serve as "containers" for data in various modalities, entailing either intrinsic properties of regions, or interrelatedness with other regions. Different types of spatial piror, e.g., spatial proximity, is also crucial, to incorporate the inherent relations entailed by geographic locations. Most studies utilize multiple data perspectives, e.g., the region embedding studies~\cite{fu2019efficient,du2019beyond,zhang2019unifying,zhou2023heterogeneous,luo2022urban,zhang2022region,li2024urban,sun2024urban} use POIs, human trajectories, and spatial proximity. The modeling of different perspectives is often intertwined, so we present each study holistically, covering the entire analytical pipeline. To this end, we categorize the studies according to the data modalities used. See Table~\ref{tab: region} for surveyed studies.

\subsection{Urban regions}
\subsubsection{Intrinsic Attributes}

Among various data modalities, POIs and imagery data, including remote sensing (RS) images and street view images (SVIs), are most commonly utilized to reflect the intrinsic properties of each region. POIs capture socioeconomic factors, while images reflect the overall physical appearance from a human or aerial perspective. Several studies have explored these data sources to learn region representations in a self-supervised manner, sometimes without much consideration of contextual information.

In a pioneering study on POI-based region representation learning, ~\cite{liu2018efficient} treats each region as an ``image" where some pixels are filled with POIs. This approach allows regions to be processed by a CNN, with POI category information, represented by one-hot encoding, serving as ``pixel values". The model is trained with the objective of a triplet loss. For each anchor region, its augmentation generated by random removal, addition, and shifting of POIs serves as a positive sample, while negative samples are non-overlapping regions or augmentations with larger perturbations (hard negative samples). UrbanCLIP\allowbreak ~\cite{yan2024urbanclip} utilizes RS images and LLM to learn region representations. For each image, it generates a detailed description using a pre-trained LLM, where detailed and specific prompts oriented to urban infrastructures are found to be beneficial. The method then trains an image encoder and a text encoder using contrastive learning and auto-regressive text generation. Building upon this work, UrbanVLP~\cite{hao2024urbanvlp} employs both SVIs and RS images. It proposes to filter the LLM-generated text by a quality metric CycleScore, to avoid low-quality text generation. This approach fuses remote sensing and street view images within each region, contrasting them with generated texts at both image and token levels. In addition, MMGR~\cite{bai2023geographic} proposes to fuse RS images and POIs for learning intrinsic attributes of urban regions. It extends the idea in~\cite{SSL_image1} by carrying out contrastive learning between multiple augmentations of each image, as well as between images and POIs using the method proposed in~\cite{semantic_poi}.

\subsubsection{Context information}

Incorporating diverse sources of context information to model the semantics of urban regions has become a standard practice. The contextual information often comes from connectivity patterns exhibited in human trajectories, spatial proximity, temporal patterns, and other types of property similarities (e.g., from land use data and knowledge graphs). The primary incentive for utilizing context information is its ability to significantly enhance the quality of learned region representations, especially when the expressiveness of data modalities representing intrinsic attributes, such as POIs, is limited. We organize the subsequent content based on the data modalities utilized to learn representations for urban regions, allowing for a focused discussion on how different types of data contribute to the modeling process.

\paratitle{POIs/Knowledge Graphs.} Based on the pioneering work~\cite{liu2018efficient}, SKRL4RS~\cite{jin2019learning} further enhances the information of spatial entities by incorporating the rich context information from two well-established knowledge graphs, YAGO and DBpedia. Instead of one-hot category encoding, the spatial entities in the knowledge graphs are transformed into representations that encapsulate the relatedness derived from their hierarchical categories (ontologies) and their proximity within the knowledge graph. In this way, the semantics of spatial entities within regions are better captured. For example, the similarity between a Japanese restaurant and a Korean restaurant is much larger than that between a factory and a Korean restaurant. 
In another research line, HGI~\cite{huang2023hgi} extends beyond the POI category 

\onecolumn
\begin{center} 
{\fontsize{8}{10}\selectfont
\begin{longtable}{   
    >{\raggedright\arraybackslash}m{2cm}
    >{\centering\arraybackslash}m{1.2cm}
    >{\centering\arraybackslash}m{1.5cm}
    >{\centering\arraybackslash}m{2.8cm}
    >{\centering\arraybackslash}m{2.8cm}
    >{\centering\arraybackslash}m{2.3cm}
    >{\centering\arraybackslash}m{3.7cm}}
     \caption{A summary of the surveyed papers on self-supervised learning for \textbf{urban regions}.} \label{tab: region}\\

    \toprule
     \textbf{Method} & \textbf{Type}  & \textbf{Data Source} & \textbf{Data Augmentation} &   \textbf{Context Information}  & \textbf{Objective Function} & \textbf{Pretext Tasks}  \\  
    \hline
    \endfirsthead

\multicolumn{7}{c}{{\bfseries Table \thetable\ A summary of the surveyed papers on self-supervised learning for \textbf{urban regions} (continued).}} \\
    \toprule
     \textbf{Method} & \textbf{Type}  & \textbf{Data Source} & \textbf{Data Augmentation} &   \textbf{Context Information}  & \textbf{Objective Function} & \textbf{Pretext Tasks}  \\  
    \hline
    \endhead

    \midrule
    \multicolumn{6}{r}{{Continued on next page}} \\ 
    \endfoot

    \bottomrule
    \endlastfoot

    Triplet\cite{liu2018efficient} & Contrastive & POI   &  POI removal, addition, and shifting & None  & Triplet loss & Region-region contrast  \\ \hline
    SKRL4RS\cite{jin2019learning} & \tabincell{c}{Predictive, \\ Contrastive} & Knowledge graph & Spatial entity shifting & Knowledge graph & Triplet loss & Region-region contrast  \\ \hline
    HGI\cite{huang2023hgi} & Contrastive & POI   & POI and region graph & \tabincell{c}{Spatial proximity,\\ City overall context} & MI maximization & \tabincell{c}{POI-region contrast,\\ Region-city contrast}  \\ \hline
    HDGE\cite{wang2017region} & Predictive & Human trajectory & Heterogeneous region graph & \tabincell{c}{Human mobility,\\ Spatial proximity,\\ Temporal pattern} & \tabincell{c}{Skipgram,\\ KL divergence} & \tabincell{c}{Reconstruction of human \\ mobility patterns}  \\ \hline
    ZE-Mob\cite{yao2018representing} & Predictive & Human trajectory & Human mobility event & \tabincell{c}{Human mobility,\\ Spatial proximity} & Skipgram & Context neighbor prediction \\ \hline
    GMEL\cite{liu2020learning} & Predictive & Human trajectory & Region graph & \tabincell{c}{Human mobility,\\ Spatial proximity} & MSE   & \tabincell{c}{Commuting flow and \\ in/out flow prediction} \\ \hline
    MGFN\cite{wu2022multi} & Predictive & Human trajectory & Region graph via clustering & Human mobility & KL divergence & \tabincell{c}{Reconstruction of human \\ mobility patterns}  \\ \hline
    MP-VN\cite{fu2019efficient} & Predictive & Human trajectory, POI & POI graph & \tabincell{c}{Human mobility,\\ Spatial proximity, \\ POi distribution similarity} & MSE   & Graph reconstruction  \\ \hline
    DLCL\cite{du2019beyond} & Predictive & Human trajectory, POI & POI and region graph & \tabincell{c}{Human mobility,\\ Spatial proximity} & Cross-Entropy & Region graph reconstruction\\ \hline
    CGAL\cite{zhang2019unifying} & Predictive & Human trajectory, POI & POI and region graph & \tabincell{c}{Human mobility,\\ Spatial proximity} & MSE, adversarial loss & \tabincell{c}{Reconstruction of \\ node feature\\ and graph structure}  \\ \hline
    HAFusion\cite{sun2024urban} & Predictive & \tabincell{c}{Human \\ trajectory,  \\POI, \\ Land use} & None  & \tabincell{c}{Human mobility, \\POIs} & MSE, KL divergence & \tabincell{c}{Reconstruction of human \\ mobility patterns, POI, \\ and land use feature \\ similarity}  \\ \hline
    MVURE\cite{zhang2021multi} & Predictive & \tabincell{c}{Human \\ trajectory, \\ POI, \\ User check-in} & Multi-view graph & \tabincell{c}{Human mobility,\\ POIs, \\ User check-in similarity} & MSE, KL divergence & \tabincell{c}{Reconstruction of human \\ mobility patterns, POI, \\ and check-in distributions} \\ \hline
    HREP\cite{zhou2023heterogeneous} & \tabincell{c}{Predictive, \\ Contrastive} & Human trajectory, POI & Heterogeneous region graph & \tabincell{c}{Human mobility,\\ Spatial proximity, \\POIs} & \tabincell{c}{KL divergence,\\ Triplet loss, MSE} & \tabincell{c}{Region-region contrast,\\ Reconstruction of human \\ mobility patterns \\ and POI distributions}  \\ \hline
    HUGAT\cite{kim2202effective} & Predictive & \tabincell{c}{Human \\ trajectory, \\ POI, \\ Land use} & \tabincell{c}{Construction of heterogeneous \\urban graph and meta-path} & \tabincell{c}{Human mobility,\\ Spatial proximity,\\ Temporal pattern} & KL divergence, MSE & \tabincell{c}{Reconstruction of human \\ mobility patterns, check-in, \\ and land use distributions} \\ \hline
    Region2Vec\cite{luo2022urban} & Predictive & Human trajectory, POI & Multi-view region graph & \tabincell{c}{Human mobility,\\ Spatial proximity, \\POIs} & KL divergence, MSE & \tabincell{c}{Reconstruction of human \\ mobility patterns,\\ geospatial adjacency\\ and POI distributions} \\ \hline
    ReMVC\cite{zhang2022region} & Contrastive & Human trajectory, POI & \tabincell{c}{POI insertion, \\deletion, and replacement\\ Trajectory heatmap perturbation} & \tabincell{c}{Human mobility, \\ POIs} & InfoNCE & Region-region contrast  \\ \hline
    ReCP\cite{li2024urban} & \tabincell{c}{Contrastive, \\ Predictive} & Human trajectory, POI & \tabincell{c}{POI insertion, \\ deletion, and replacement\\ Trajectory heatmap perturbation} & \tabincell{c}{Human mobility, \\ POIs} & \tabincell{c}{InfoNCE,\\ MI maximization,\\ Conditional entropy} & \tabincell{c}{Region-region contrast, \\ Region feature reconstruction} \\ \hline
    Tile2Vec\cite{jean2019tile2vec} & Contrastive & RS    & None  & Spatial proximity & Triplet & Region-region contrast  \\ \hline
    RegionEncoder\cite{jenkins2019unsupervised} & Predictive &  \tabincell{c}{RS,\\ Human \\ trajectory, \\ POI} & Image noising, region graph & Human mobility & \tabincell{c}{MSE,\\ KL divergence,\\ Cross-Entropy} & \tabincell{c}{Reconstruction of images \\ and human mobility \\ patterns , Multi-view \\ graph discrimination} \\ \hline
    Urban2Vec\cite{wang2020urban2vec} & Contrastive & SVI, POI & None  & Spatial proximity & Triplet & Image- and region-level contrast \\ \hline
    M3G\cite{huang2021learning} & Contrastive & \tabincell{c}{Human \\trajectory, \\POI, \\ SVI} & None  & Spatial proximity & Triplet & \tabincell{c}{Region-SVI,\\ region-POI,\\ region-region contrast} \\ \hline
    Xi et al\cite{xi2022beyond} & Contrastive & RS, POI & None  & Spatial proximity & NT\_Xent & Region-region contrast  \\ \hline
    UrbanVLP\cite{hao2024urbanvlp} & Contrastive & RS, SVI & Textual descrptions of images & None  & InfoNCE & \tabincell{c}{Image- and token-level\\ Image-text alignment} \\ \hline
    MMGR\cite{bai2023geographic} & Contrastive & RS, POI & RS image augmentation, POI graph & None  & InfoNCE & Region-region contrast \\ \hline
    UrbanCLIP\cite{yan2024urbanclip} & \tabincell{c}{Predictive, \\ Contrastive} & RS    & Textual descrptions of images & None  & \tabincell{c}{InfoNCE,\\ Cross-Entropy} & \tabincell{c}{Image-text alignment,\\ Text autoregressive \\ prediction}  \\ \hline
    GeoHG\cite{zou2024learning} & Contrastive & RS, POI & Heterogenous graph & Spatial adjacency, Region feature similarity  & InfoNCE & Graph contrast \\ \hline
    ReFound\cite{refound} & \tabincell{c}{Predictive, \\ Contrastive} & RS, POI & \tabincell{c}{Foundation model distillation\\ and text generation} & None  & \tabincell{c}{KL divergence\\ Cross-entropy} & \tabincell{c}{Knowledge distillation\\ Masked data modeling \\ Cross-modal spatial alignment} \\ \hline
    MuseCL\cite{musecl} & Contrastive & RS, SVI, POI & None & Region feature similarity  & \tabincell{c}{Triplet\\ InfoNCE} & \tabincell{c}{Image-level contrast\\ Image-text contrast} \\ \hline
    GeoHG\cite{zou2024learning} & Contrastive & RS, POI & Heterogenous graph & Spatial adjacency, region feature similarity  & InfoNCE & Graph contrast \\ \hline 
    RegionDCL\cite{li2023urban} & Contrastive & \tabincell{c}{OSM building \\footprints, \\ POI} & \tabincell{c}{Point injection, \\ Building removal} & Spatial proximity & \tabincell{c}{InfoNCE, \\ Triplet loss} & Region-region contrast
    
\end{longtable}}
\end{center}
\twocolumn

\noindent embedding technique in~\cite{semantic_poi} by further applying GNN aggregation process to escalate the representations to region and city levels. This model is optimized by maximizing the mutual information among the POI-region-city hierarchy.

\paratitle{Human Trajectories.} Human mobility data, i.e., trajectories, has been a popular data source for learning region representations, mainly in merit of the rich region-level connectivity patterns exhibited from massive trajectories. HDGE~\cite{wang2017region} is the first to leverage such data sources to consider both temporal dynamics and multi-hop transition patterns between regions. Specifically, it defines a flow graph where each node represents a region at a certain time point. Nodes in the flow graph are connected by two types of edges based on human flow between regions and spatial adjacency. The spatial adjacency edges are built to mitigate the data sparsity problem in human trajectories, i.e., to gauge the location when a user’s location is not recorded. The model is trained to reconstruct the transition probability between regions, i.e., minimizing the KL-divergence between the skip-gram probability in the graph and the empirical transition probability observed from the trajectory dataset.

Later, ZE-Mob~\cite{yao2018representing} defines several human mobility events pertaining to regions, time, and movement mode (i.e., departure/arrival). In this way, region representations can be learned similarly in Word2vec~\cite{word2vec} by skip-gram objective. Besides, it enriches the objective by integrating the importance of region origin-destination pairs based on popularity and distance. GMEL~\cite{liu2020learning} constructs a region adjacency graph and subsequently employs two graph attention networks to model the two types of representations for regions functioning as travel origins and destinations. The model is trained with the pretext tasks of predicting human flow and predicting in/out flow. MGFN~\cite{wu2022multi} regards human movements in each time slot as a mobility graph and defines several graph distance measures to cluster the graphs into several mobility patterns, e.g., distance between mean or variance of edge weights and human flow imbalance. A hierarchical clustering method is then applied to distill these into a reduced number of region graphs that represent specific mobility patterns. Within each mobility pattern graph, message passing is performed. Besides, inter-pattern attention mechanisms are employed to fuse various mobility patterns for the final region representations. The model is trained with the objective of reconstructing region-level human mobility patterns.

\paratitle{POIs + Human Trajectories (+others).} Many region representation studies integrate POIs with human trajectories, as these data sources naturally complement each other. POIs describe the range of activities within a region, while human trajectories reveal inter-regional connections. This integration is often enhanced with additional data, such as land use. For instance, MP-VN~\cite{fu2019efficient} constructs two POI graphs to capture both static and mobility patterns among POI types, based on mobility connectivity and geographic distances. These graphs 
are sent to an autoencoder for learning region representations via graph reconstruction, while also incorporating spatial proximity and functional similarity derived from POIs. Building on this, DLCL~\cite{du2019beyond} employs an adversarial autoencoder to learn from these two POI graphs. CGAL~\cite{zhang2019unifying} enhances this model by introducing collective adversarial learning, using an assemble-disassemble strategy where fused region representations are disaggregated to reconstruct the original graphs. It captures region similarities based on POI distributions, textual information, and temporal patterns from human trajectories. HAFusion~\cite{sun2024urban} further utilizes human trajectories, POIs, and land use data to model urban regions. Each region is characterized by mobility features, POI categories, and land use counts, with an attention-based encoder capturing intra- and inter-view region correlations, while a fusion module integrates multi-view embeddings and captures higher-order correlations.

Another line of research extensively employs GNNs.  MVURE \allowbreak ~\cite{zhang2021multi} uses human trajectories, POIs, and check-ins to construct four graph views, encoding each view with a graph attention network. It implements cross-view information sharing through attention mechanisms and fuses the views using adaptive weighting, aiming to reconstruct mobility patterns and region relatedness from POIs and check-ins. HREP~\cite{zhou2023heterogeneous} enhances this by defining multiple edge types from human mobility, POIs, and geographic context, with source and target edges based on trajectories, POI similarities, and geographic adjacency. It learns region representations using an attention-based GNN and three objectives: geographic proximity loss, mobility reconstruction, and POI correlation reconstruction. Additionally, HREP uses prompt learning (prefix-tuning) to adapt learned representations for downstream tasks, a method borrowed from NLP. In addition, HUGAT~\cite{kim2202effective} utilizes POI check-in trajectories and land use data for region representation learning. It defines five types of nodes, e.g., regions and POI categories, and two types of edges for spatial and temporal relations. This method then constructs a heterogeneous information network and designs five meta-paths for interesting relations like regions that are popular destinations at the same time. A heterogeneous graph attention network is employed to derive region representations by minimizing the difference between estimated and actual mobility patterns of regions, check-in distributions, and land use distribution. Region2Vec~\cite{luo2022urban} constructs a multi-graph using human trajectories, spatial adjacency, and POI distribution embeddings from a knowledge graph. GNN and attention-based fusion layers are employed to create region representations, with objectives of reconstructing mobility patterns, maintaining spatial adjacency, and preserving POI distribution similarities.

The third line employing POIs and human trajectories focuses on the application of the contrastive learning paradigm. ReMVC~\cite{zhang2022region} is a dual-view approach integrating POIs and human mobility data. For the POI view, it uses region-level POI category proportions as raw features, and performs data augmentation through random POI insertion, deletion, and replacement. These augmented POI views of regions serve as positive samples, whereas POI data from differing regions are used as negative samples for the contrastive learning. For the human mobility view, it constructs two heatmaps to represent the source and destination patterns of each region, followed by the augmentation of Gaussian noise injection to form positive samples for the contrastive learning. Furthermore, an inter-view contrastive learning objective is employed to ensure that representations of a region from different perspectives are similar, while the representations of different regions are distinguishable. ReCP~\cite{li2024urban}  develops a fusion technique for multiple information views from an information theory perspective. Apart from maximizing the mutual information (consistency) shared between different views, this method implements a dual prediction strategy to minimize the conditional entropy between representations from different views, thereby reducing the inconsistency between views.

\paratitle{Imagery Data}. Imagery data, including RS images and SVIs, have long been established to represent the physical appearance of urban environments from both aerial and ground-level human perspectives~\cite{patino2013review,fan2023urban}. Such visual appearances of urban environments can be used to partially reflect the socioeconomic factors in cities. Therefore, it has become increasingly popular to utilize imagery data for learning region representations, often in conjunction with additional data sources and context information, such as spatial proximity.

Tile2Vec~\cite{jean2019tile2vec} generates representations based on RS images patches for square-shaped regions. It employs a CNN model to encode remote sensing images and a triplet loss to ensure that patches which are geographically adjacent are also close in the embedding space. RegionEncoder~\cite{jenkins2019unsupervised} leverages RS images, POIs, and human trajectories for learning region representations. Initially, it employs a denoising autoencoder to extract representations from RS images. Following this, a region graph is constructed, utilizing region-level POI features as node attributes and human trajectory data to define inter-region connectivity through edges. The model is trained with two reconstruction losses, and a cross-modal alignment objective. The study in ~\cite{xi2022beyond} adopts a contrastive learning strategy to maximize the similarity of representations derived from adjacent RS images and those that have similar POI distributions. The representations learned from the two pathways of each region are adaptively fused using learnable weights in downstream tasks.

Urban2Vec~\cite{wang2020urban2vec} derives the initial region representations by averaging all the SVI representations obtained from an image encoder for each region. It subsequently fine-tunes this image encoder with a spatial proximity-based triplet loss to bring SVIs that are spatially close together in the embedding space. Besides, this method integrates the POIs by further generating the POI representation of a region that encapsulate all the words associated with the POIs in each region. It introduces another triplet loss designed to merge POI information into the region representations by minimizing the distance between each region's representation and its corresponding POI representation.
M3G~\cite{huang2021learning} extends Urban2Vec by simultaneously utilizing SVIs, POIs, and human trajectories. M3G enhances the model by conducting region-level contrastive learning, selecting other regions that are either spatially close or connected through human mobility as positive samples. In another work, GeoHG~\cite{zou2024learning} used the semantic segmentation features of the European Space Agency images system for processing remote sensing images, and constructed a heterogeneous graph to capture the spatial adjacency and similarities at the region level reflected from the similarities in environmental and socioeconomic characteristics. The graph model was pretrained using contrastive learning.

Another interesting region embedding study using both street view and satellite images, in conjunction with POIs and human trajectories is MuseCL~\cite{musecl}. The method carries out contrastive learning at multiple levels. First, contrast was conducted for street view images based on travel pattern similarity reflected form human trajectories. Likewise, contrast was carried out for remote sensing images based on POI similarity. The rationale of such pairing is that street view and mobility data entail human movement patterns, while remote sensing and POI data both capture the built environment and land use. The final region embeddings are obtained through fusing street view and remote sensing features, and supplemented with textual information from POIs. 

With the rapid advancements in vision foundation models that are pretrained with vast amounts of images or image-text pairs, leveraging the power of foundation models for region embedding has become a trend. To this end, ~\cite{refound} proposed ReFound with POIs and remote sensing images. One major argument here is the distillation of knowledge contained in foundation models is beneficial. To this end, the POI embeddings are enforced to be close to the embeddings of LLM-generated text, image embeddings are pushed towards the representations generated by vision foundation models, and the similarities between image-POI pairs also imitate the similarities derived from a vision-language model. ReFound is finally pretrained by combining distillation objectives, masked data modeling, and cross-modal modality alignment.

\paratitle{OpenStreetMap.} A new trend in region representation is to utilize data from OpenStreetMap (OSM)~\cite{osm}, a global-scale open geospatial dataset contributed by a community of mappers. OSM offers a extensive repository of geospatial entities such as building footprints and road networks, serving as valuable and accessible resources for learning effective region representations. RegionDCL~\cite{li2023urban} extracts building footprints and POIs from OSM, and begins by encoding the shape information from building footprints using a CNN-based model to generate initial region features. Furthermore, RegionDCL addresses the challenge of empty areas, which are spaces not explicitly represented in discrete geospatial vector data but are physically present. To effectively represent these empty areas in the final region embeddings, this method employs Poisson Disk Sampling to fill these gaps. Each building group--small regions partitioned by road networks--is then processed through a distance-biased Transformer. This Transformer is trained using building group-level contrastive learning. Subsequently, these building groups are aggregated into larger regions for a second round of contrastive learning, employing a triplet loss with an adaptive margin to refine the final region embeddings.

\subsubsection{Applications}

As urban regions serve as a critical analytical scale for various urban analyses and prediction tasks, the learned region representations are used in a diverse range of downstream tasks. Commonly, these region embeddings are utilized by integrating them as frozen inputs into shallow task predictors, such as MLP, for making task-specific inferences. Additionally, several studies have enhanced the utilization of region representations in downstream tasks through advanced methods like prompt learning~\cite{zhou2023heterogeneous} and adaptive multi-view fusion~\cite{xi2022beyond}. From the perspective of downstream tasks, the predominant tasks involve predicting various socioeconomic indicators in cities. These include region-level attributes such as land use/urban function/land cover~\cite{yao2018representing, zhang2022region, li2024urban}, population density~\cite{wang2020urban2vec, huang2023hgi}, house prices~\cite{jenkins2019unsupervised, yan2024urbanclip}, average income~\cite{huang2021learning, wang2017region}, check-in counts\allowbreak ~\cite{fu2019efficient, zhang2019unifying}, crime rates~\cite{wu2022multi, wang2017region}, service call volumes~\cite{sun2024urban}, GDP~\cite{bai2023geographic, yan2024urbanclip}, nighttime lighting~\cite{hao2024urbanvlp}, takeaway order volumes~\cite{xi2022beyond}, health indices~\cite{jean2019tile2vec}, and so on. Moreover, region representations are extensively utilized for tasks like similar region search\cite{liu2018efficient, jin2019learning} and region or land use clustering~\cite{luo2022urban} in a fully unsupervised manner, which has significant practical implications for real-world urban planning and management. Furthermore, region representations can benefit studies on human mobility in cities, such as predicting human flow and bike flow~\cite{liu2020learning, kim2202effective}. Overall, the application of SSL-based region representations has proven to be diverse and effective across many critical urban analytical tasks, leading to substantial real-world benefits. As urban environments continue to expand and evolve, the role of robust region representations in tackling complex urban challenges will become increasingly crucial, paving the way for smarter, more resilient cities.

\section{Multi-type Learning}\label{sec:multi-type}
While SSL techniques have demonstrated promising performance in producing representations for geospatial objects, these methods focus on separately deriving representations for individual geospatial data types. This paradigm does not consider the complex interactions and potential synergies among various geospatial data types. To this end, recent studies propose multi-type learning methods, which involve the joint modeling for multiple geospatial data types, as a step towards the development of  geospatial foundation models. 

One research direction involves the construction of geospatial knowledge graphs  to systematically represent various geospatial objects as nodes and their relationships with diverse data sources as edges. By structuring geospatial entities within a graph-based framework, knowledge graphs enable the encoding of complex spatial, semantic, and contextual relationships among various data types. Once constructed, graph embedding techniques can be applied to derive representations that capture both the intrinsic properties of geospatial objects and their interactions within the broader spatial environment. WorldKG\cite{worldkg} aligns geospatial objects from OpenStreetMap with ontologies in existing knowledge graphs such as Wikidata. It structures geospatial knowledge by defining graph nodes based on geospatial objects, semantic classes, and object attributes. In contrast, frameworks such as UUKG~\cite{uukg}, UrbanKG~\cite{urbankg}, and UrbanFlood~\cite{urbanflood} develop customized graph schemas tailored to specific research perspectives, such as user aspect~\cite{urbankg}, hierarchical spatial resolution~\cite{uukg}, or application-oriented~\cite{urbanflood} geospatial analysis. Following the construction of these knowledge graphs, various graph embedding techniques~\cite{graph_embedding_survey} can be employed to generate representations for each node. These embeddings facilitate downstream geospatial tasks by leveraging the rich structural and semantic information encoded within the knowledge graph framework from multiple data types.

Besides, other studies formulate this problem through a heterogeneous learning framework using contrastive SSL. HOME-GCL~\cite{home-gcl} aims to derive representations for both road segments and regions. Specifically, it constructs a heterogeneous graph  that incorporates multi-view intra-entity relationships based on geographical distance, functionality, and human mobility records, as well as inter-entity connections based on topological containment. A heterogeneous GNN is then applied to aggregate features among different entities. Subsequently, intra-level contrastive learning is employed to distinguish entities from the same object after graph augmentation,  while inter-level contrastive learning differentiates between connected and disparate object. CityFM~\cite{cityfm} simultaneously considers three data types. It encodes the textual information for each entity from these data types using pre-trained language models and employs contrastive learning to contrast an entity to its spatial neighbors. Besides, it incorporates visual content from regions, applying visual encoders to derive corresponding representations, which are then contrasted with textual representations. Lastly, CityFM utilizes contrastive learning to encourage road segments with similar traffic patterns to develop similar representations. For multi-type learning studies, the derived representations encode complementary interactions from multiple data types. As a result, these representations are effectively utilized in downstream applications, particularly those involving multiple data types, such as site selection.
\section{Discussion}\label{sec:trending}
Sections~\ref{sec:points}-\ref{sec:multi-type} have presented specialized SSL techniques designed for different geospatial data types. Through an extensive review of prior studies, we observe a notable evolution in SSL methodologies for geospatial representation learning, characterized by the following key trends:

\begin{enumerate}[leftmargin=*]
    \item  \textbf{Transition from Single-Source to Multi-Source Context Integration}: Early research predominantly derived representations from a single source of context information. However, recent advancements increasingly emphasize multi-source and multi-modal integration, leveraging diverse and complementary context signals to better capture intrinsic spatial patterns in geospatial data.
    \item \textbf{Transition from Naive Models to Hybrid Architectures}: 
    Initial approaches typically employ simple models (e.g., Word-\allowbreak 2vec or RNN variants) for geospatial representation learning. Over time, research has shifted towards hybrid models that combine multiple and complex architectural components, enabling more expressive modeling of geospatial context information.
    \item \textbf{Influence of Foundation Models from Visual and Language Domains}: The rapid progress in visual and language foundation models has significantly influenced geospatial representation learning. Recent studies increasingly incorporate these foundation models as integral components or explore methods to align the implementation of these pre-trained models in the context of geospatial data.
\end{enumerate}

While the studies belonging to the first two research trends have been well-established and detailed in preceding sections, the emergence of geospatial foundation models remains an evolving area with no fixed formulation and is explored from various perspectives. Recognizing this new emerging trend, we provide an overview of recent efforts in developing geospatial foundation models from multiple perspectives.

Additionally, while the primary focus of this survey is to summarize SSL-based studies that adopt a task-agnostic approach to geospatial representation learning, another important research direction involves task-specific SSL techniques. Unlike general-purpose SSL, these methods employ self-supervised objectives as auxiliary learning signals to enhance supervised tasks in geospatial applications. To offer a broader perspective, we also present a brief overview of research efforts that incorporate task-specific SSL into several typical geospatial applications.

\subsection{A Step Towards Geospatial Foundation Models}\label{subsec:LLM}

Foundation models, which are fundamentally rooted in the SSL paradigm with pretext tasks such as autoregressive modeling in language domains and masked patch recovery in vision domains, have emerged as a transformative trend across various domains, including GeoAI. Foundation models have demonstrated significant advancements in understanding, reasoning, and generative capabilities across diverse data modalities, including language~\cite{GPT4}, images~\cite{CLIP, CoCA}, graphs~\cite{LLMgraph1, LLMgraph2}, and time series~\cite{LLMtimeseries1, LLMtimeseries2}. Inspired by the success, researchers have increasingly explored the development of geospatial foundation models primarily based on language models from various perspectives.

One perspective involves enhancing existing foundation models with geospatial knowledge. Empirical studies have shown that LLMs, such as the GPT series, already possess a certain level of geospatial knowledge~\cite{evaluate-llm-survey, llm-geoknowledge1, llm-geoknowledge2}.
To further \allowbreak strengthen their geospatial knowledge understanding, \allowbreak researchers have proposed the development of geospatial language models~\cite{k2, geolm, bbgeogpt} by fine-tuning general-purpose LLMs within an autoregressive SSL paradigm. These models are trained on domain-specific corpora, including research publications and Wikipedia pages related to geography, urban planning, and spatial sciences, enabling them to perform better on geospatial tasks such as geospatial question answering. 

Beyond continued fine-tuning, another perspective explores leveraging LLMs for complex geospatial problem-solving \allowbreak through planning, decision-making, and pipeline orchestration. LLMs are increasingly utilized as intelligent agents capable of organizing systematic workflows for geospatial tasks. This is \allowbreak achieved through task decomposition, strategic tool selection, and model integration, where LLMs interact with external spatiotemporal models and analytical tools to process queries and generate informed responses. Notable examples include UrbanLLM~\cite{urbanllm}, GeoGPT~\cite{geogpt}, PlanGPT~\cite{plangpt}, and MapGPT~\cite{mapgpt}, which are designed to answer geospatial queries by organizing the workflow of solving these queries by calling external functions (e.g., ArcGIS) and integrating outputs from them. While these approaches contribute to the broader development of geospatial foundation models, they are often not explicitly designed for direct representation learning of geospatial objects.

Moreover, several studies have explored to enhance the interaction between LLMs and particular geospatial objects through targeted alignment strategies by SSL.
For example, LAMP~\cite{lamp} aims to infuse fine-grained knowledge about POI into LLMs for a specific city, subsequently facilitating POI-related applications in a conversational manner. To achieve this, it structures several POI search tasks with their ground truth data as templates within the SSL corpus to fine-tune the LLM. Consequently, LAMP can solve several POI applications, such as route recommendation and location search, by formulating them as query-response processes within LLM framework. For polyline object, TrajFM~\cite{trajfm} develops a trajectory modeling framework that pre-trains LLMs from scratch in an autoregressive manner, integrating spatial, temporal, and POI modalities for each trajectory data point. To enhance task transferability, TrajFM employs a trajectory masking and recovery scheme, which unifies various trajectory-related task generation processes by masking and reconstructing trajectory sub-segments and modalities. This enables the pre-trained model to generalize across diverse trajectory-based tasks. TrajCogn~\cite{trajcogn} transforms trajectory features into structured, natural language-like inputs by mapping these features into descriptive words, allowing LLMs to process trajectory data effectively. The model is then fine-tuned using a cross-reconstruction pretext task, enabling generalizable utility for downstream applications such as travel time estimation, destination prediction, and trajectory similarity search. For polygon object,
GeoLLM~\cite{geollm} proposes harness the capabilities of LLMs to encode geospatial knowledge and capture regional features effectively. Specifically, it pinpoints targeted region coordinates on the map, extracts the corresponding address that contains place names from the neighborhood level up to the country, and identifies the ten nearest places. The regional coordinates with its geospatial context are then formatted into templated textual prompts as input to the LLM. After processing the prompts, the LLM is designed to be automatically fine-tuned using response variables, such as various socio-economic indicators. UrbanGPT~\cite{urbangpt} introduces a SSL instruction-tuning paradigm that seeks to align the dependencies of time and space, with the language space of LLMs. This method constructs prompts that combine  textual descriptions with representations obtained from spatio-temporal learning models for LLM. The resulting output representations encapsulate both semantic information and relevant time-space dependencies for geospatial regions.

\subsection{Task-specific SSL techniques}

As previously discussed, task-specific SSL techniques serve as auxiliary objectives to enhance performance in specific geospatial tasks. These methods are particularly relevant for data instances such as grids (rasters) and point sensors, where various types of measurements (e.g., flows, speeds) are recorded over time, forming a collection of time series from these instances that exhibit both spatial and temporal interdependencies. 

Therefore, a notable application of task-specific SSL applied to these data instances is in the application of spatio-temporal forecasting based on the collected records. For example, models like UniST~\cite{unist}, W-MAE~\cite{w-mae}, and GPT-ST~\cite{gpt-st} are utilized for grids or sensor points, employing a strategy of reconstructing masked features as self-supervised pre-training method to learn dynamic dependencies. The learned parameters are then fine-tuned for specific spatio-temporal forecasting datasets. Besides, SSL techniques are also integrated within multi-task learning frameworks for spatio-temporal forecasting. UrbanSTC~\cite{urbanstc} applies contrastive learning to identify grid regions in both spatial and temporal dimension with similar patterns. STGCL~\cite{stgcl} explores various contrastive learning schemes with in the framework of spatio-temporal GNN, such as node-level and graph-level contrasts, providing some insights into effective integration strategies. ST-SSL~\cite{st-ssl} performs the adaptive augmentation over the traffic flow graph at both attribute- and structure-levels within the spatio-temporal GNN framework. It introduces two SSL auxiliary tasks to supplement the main traffic forecasting task to account for spatial and temporal heterogeneity.  SSTBAN~\cite{sstban} incorporates a masked auto-encoder module to reconstruct masked spatio-temporal patches, thus deriving more robust representations to support the forecasting task. CL4ST~\cite{cl4st} develops a meta view generator to automatically construct node and edge augmentation views for contrastive learning in a data-driven manner. Moreover, several studies also incorporate contrastive SSL into the scenario of POI recommendation~\cite{poi-ssl1, poi-ssl2, poi-ssl3}. These diverse methods demonstrate the versatility and potential of task-specific SSL techniques in enhancing the performance for various downstream applications regarding geospatial data.

\section{Future Research Directions}\label{sec:discussion}
In this section, we identify critical problems of existing SSL methods for geospatial objects, and outline several promising research directions for future exploration in this domain.

\paratitle{Selection of pretext tasks and data augmentation}.
Pretext tasks and data augmentation techniques play a crucial role in the effectiveness of SSL. Existing SSL methods for geospatial objects usually draw inspiration from the domains of computer vision and graph learning, employing heuristic adaptations tailored to geospatial contexts.  As a result, the selection of pretext tasks and data augmentation strategies can vary significantly across different geospatial SSL implementations. While there have been efforts to systematically assess the effectiveness of different pretext tasks and data augmentation techniques in other domains~\cite{eval-ssl1, eval-ssl2, eval-ssl3}, these findings may not directly translate to geospatial SSL due to the unique characteristics and diversity of geospatial data types. Therefore, there is a pressing need to investigate the impact of different pretext task and data augmentation selections specifically within the geospatial domain. Future research should explore whether certain pretext tasks or augmentation techniques offer distinct advantages for geospatial representation learning. This inquiry should ideally be guided by theoretical analyses and comprehensive empirical evaluations to ensure the robustness and generalizability of SSL techniques for geospatial objects.

\paratitle{Benchmarking SSL for geospatial objects}.
The establishment of standardized benchmarks is essential for advancing SSL techniques in geospatial domain. Unlike domains such as computer vision or natural language processing, where benchmark datasets are widely available and standardized, SSL methods for geospatial objects are currently evaluated on datasets that are often task-specific and unique to individual studies. The absence of unified benchmarking datasets hinders consistent and reliable comparisons across different methods, making it difficult to systematically assess their effectiveness and generalizability. Future research should prioritize the development of well-structured benchmarking frameworks that facilitate comprehensive evaluations of SSL models for geospatial objects. This requires a concerted effort to curate and release open-access benchmark datasets~\cite{benchmark} that encompass diverse geospatial data types, heterogeneous contextual information, and multiple downstream tasks.  Additionally, the establishment of automated benchmarking platforms would significantly advance this domain by enabling standardized evaluations of SSL models. Such platforms could provide pre-configured training  pipe-lines, fair comparison protocols, and leaderboards, similar to benchmark suites in other AI disciplines.

\paratitle{Enhanced multi-modality fusion}.
With the rapid expansion of location-based services and spatial crowdsourcing, data from diverse sources attached to geospatial objects, previously difficult to access, becomes increasingly available, including street view images, textual comments, and videos. Several efforts have been made to integrate multi-modality models to enhance the performance of geospatial applications~\cite{weiming1, GeoCLIP, AddressCLIP}. However, there remains significant untapped potential in advancing model capabilities for geospatial tasks.  Integrating multi-modality data sources for geospatial objects presents both challenges and opportunities.  Models designed for multi-modality data fusion must handle the discrepancies in scale, resolution, and relevance across different data sources. Future research can explore novel architectures, pretext tasks, and fusion techniques that  effectively leverage the complementary information from different data sources. For example, adapting models like CLIP~\cite{CLIP} with geospatial awareness.  It would also  be interesting to study how to balance the contributions of different modalities, aiming to achieve more generalizable and robust representations in real-world scenarios.

\paratitle{Geospatial foundation models and LLM adaptation}.
As discussed in Section~\ref{sec:trending}, LLMs and multi-type pre-trained models opens up exciting possibilities for adaptation to the geospatial context, serving as a universal basis for various geospatial applications. In terms of pathways in pre-training geospatial foundation models, future research could focus on creating  SSL techniques that can be efficiently trained on massive-scale geospatial datasets~\cite{massiveSSL1, massiveSSL2}. These models could learn generalized representations of geospatial features, patterns, and relationships, forming a foundation to be fine-tuned for specific tasks. On the other hand, adapting LLMs to the geospatial context requires novel learning paradigms that align spatial relationships and geographic context within the language space. For instance, there is a critical need to develop cross-modal training techniques that effectively bridge textual, visual, and spatial data for enhancing the applicability in more diverse scenarios and tasks in multiple modalities.

\paratitle{Privacy and vulnerability}.
The inherent nature of geospatial data, which often contains sensitive information about individuals and urban infrastructure, raises significant privacy and data vulnerability concerns~\cite{privacy_geoAI_vision_paper}. For instance, aggregated trajectories, when analyzed with certain approaches, can expose individual's routes and private locations~\cite{privacy1}. To this end, privacy-preserving techniques, such as differential privacy, could be considered within the SSL framework, especially the pre-training corpus. Moreover,  federated learning approaches~\cite{federated1, federated2} can be explored, which enable the collaborative training of models on sensitive geospatial data while avoiding the direct sharing of raw data, thus maintaining privacy. 
Furthermore, geospatial SSL models, like other machine learning models, are susceptible to adversarial attacks where input data is manipulated to induce model errors, which are particularly problematic in urban decision-making process. Therefore, robust SSL techniques can be developed to resist such data poisoning attacks or adversarial examples~\cite{attack1, attack2}.

\section{Conclusion}\label{sec:conclusion}

This paper provides a comprehensive overview of self-super-vised learning for geospatial objects in the domain of GeoAI. We develop a structured framework and introduce a systematic taxonomy that organizes SSL based on three geospatial data types and two methodology categories. For each data type, we offer detailed descriptions of methods, summarize their key features within the SSL component, and discuss their downstream applications, along with the utilization of multi-type SSL techniques. We further present studies on the emerging trends and task-specific SSL techniques. Finally, we outline several promising directions for the research in the future. As this domain continues to expand and evolve, we hope that the discussion presented in this paper will contribute to the future advancements of GeoAI.

\bibliographystyle{elsarticle-num} 
\bibliography{ref}






\end{document}